\pdfoutput=1

\documentclass[11pt, dvipsnames]{article}

\usepackage[preprint]{acl}

\usepackage{times}
\usepackage{latexsym}

\usepackage{pgfplots}
\usepackage{subcaption}
\usepackage{tikz}
\usepackage{scalerel}
\usepackage[T1]{fontenc}
\usepackage{algorithm}
\usepackage{algpseudocode}

\usepackage[utf8]{inputenc}

\usepackage{microtype}

\usepackage{inconsolata}
\usepackage{pgfplots}
\usepackage{subcaption} 
\pgfplotsset{compat=1.18} 

\definecolor{brandeisblue}{rgb}{0.0, 0.45, 1.0}
\definecolor{cyan}{rgb}{0.0, 1.0, 1.0}

\usepackage{hyperref}       
\usepackage{url}            
\usepackage{booktabs}       
\usepackage{amsfonts}       
\usepackage{nicefrac}       
\usepackage{microtype}      

\usepackage{graphicx}
\usepackage{amsmath}
\usepackage{amsthm}
\usepackage{placeins}
\usepackage{thmtools}
\usepackage{thm-restate}
\usepackage{amssymb}
\usepackage{enumitem}
\usepackage{wrapfig}
\usepackage{multirow}

\usepackage{cleveref}

\usepackage{algpseudocodex}
\usepackage[most]{tcolorbox}
\usepackage{colortbl}  
\usepackage{graphicx}

\usepackage{amsmath}
\usepackage{amssymb}
\usepackage{amsthm} 
\usepackage{nicefrac}
\usepackage{url}
\usepackage{booktabs}

\declaretheorem[name=Definition,numberwithin=section]{definition}

\newcommand{\mi}[3][]{I_{#1}(#2;#3)}

\newcommand{\unlearnedcolor}{\color{SkyBlue}}
\newcommand{\residualcolor}{\color{Dandelion}}

\newcommand{\unlearned}{{\unlearnedcolor \textbf{unlearned}}}
\newcommand{\Unlearned}{{\unlearnedcolor \textbf{Unlearned}}}
\newcommand{\residual}{{\residualcolor \textbf{residual}}}
\newcommand{\Residual}{{\residualcolor \textbf{Residual}}}
\newcommand{\unlearnedmath}{{$\unlearnedcolor{I^B_{\text{uniq}}}$}}
\newcommand{\residualmath}{{$\residualcolor{I_\cap}$}}

\definecolor{green1}{HTML}{d3e29d}
\definecolor{green2}{HTML}{acc864}
\definecolor{green3}{HTML}{8ab446}
\definecolor{green4}{HTML}{2d6b22}
\definecolor{myred}{HTML}{EFB9AD}

\tcbset{
  takeaway/.style={
    colback=blue!5!white,  
    colframe=blue!75!black,  
    boxrule=0.5pt,
    arc=2pt,
    left=6pt,
    right=6pt,
    top=4pt,
    bottom=4pt,
    fonttitle=\bfseries,
    before skip=8pt,
    after skip=8pt,
    enhanced,
    sharp corners,
  }
}

\title{Auditing Language Model Unlearning via Information Decomposition}

\author{Anmol Goel$^{\xi,\psi}$ \quad Alan Ritter$^\zeta$ \quad Iryna Gurevych$^{\xi,\psi}$ \\ 
$^\xi$Ubiquitous Knowledge Processing Lab (UKP Lab), \\
Department of Computer Science, Technical University of Darmstadt \\ 
$^\psi$National Research Center for Applied Cybersecurity ATHENE, Germany\\
$^\zeta$College of Computing, Georgia Institute of Technology \\
\url{http://www.ukp.tu-darmstadt.de/}
}

\begin{document}
\maketitle

\begin{abstract}
We expose a critical limitation in current approaches to machine unlearning in language models: despite the apparent success of unlearning algorithms, information about the forgotten data remains linearly decodable from internal representations. To systematically assess this discrepancy, we introduce an interpretable, information-theoretic framework for auditing unlearning using Partial Information Decomposition (PID). By comparing model representations before and after unlearning, we decompose the mutual information with the forgotten data into distinct components, formalizing the notions of \unlearned{} and \residual{} knowledge. Our analysis reveals that redundant information, shared across both models, constitutes \residual{} knowledge that persists post-unlearning and correlates with susceptibility to known adversarial reconstruction attacks. Leveraging these insights, we propose a representation-based risk score that can guide abstention on sensitive inputs at inference time, providing a practical mechanism to mitigate privacy leakage. Our work introduces a principled, representation-level audit for unlearning, offering theoretical insight and actionable tools for safer deployment of language models.\footnote{Code: \href{https://github.com/UKPLab/eacl2026-auditing-unlearning}{https://github.com/UKPLab/eacl2026-auditing-unlearning}}

\end{abstract}

\section{Introduction}
The rapid adoption of Large Language Models (LLMs) introduces pressing challenges around user data, including privacy \citep{das2025security} and copyright concerns \citep{karamolegkou2023copyright}. Studies have shown that LLMs often memorize sensitive information like personally identifiable data \citep{menta2025analyzing,yao2024machine,zhang2024safe,patil2023can,niu2024does,goel2025differentially}. Although users may initially consent to data usage, they may later revoke that consent. Legal frameworks like the GDPR 
\footnote{\url{https://gdpr-info.eu/}}
and the EU AI Act 
\footnote{\url{https://artificialintelligenceact.eu/}}
enforce the ``right to be forgotten,'' mandating the ability to remove user data from machine learning systems. This has led to growing interest in machine unlearning methods that aim to eliminate the influence of specific data samples from trained models \citep{bourtoule2021machine}.
\begin{figure}[!t]
    \centering
    \includegraphics[width=\linewidth]{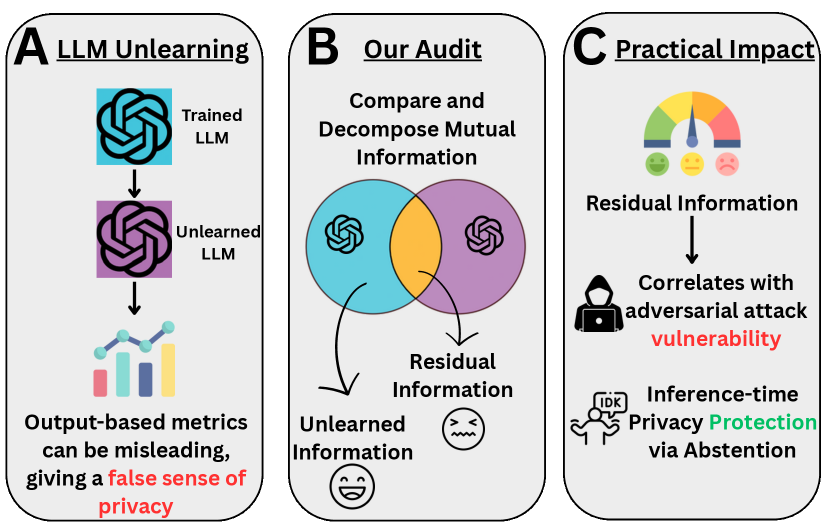}
    \caption{Existing unlearning methods may appear effective but retain forgotten data internally (A). Our framework (B) uses Partial Information Decomposition to reveal \residual{} knowledge, which (C) correlates with attack risk and informs a privacy-aware abstention score.}
    \vspace{-20pt}
    \label{fig:motivation}
\end{figure}
Existing unlearning techniques can be broadly taxonomized as \emph{exact} or \emph{approximate} \citep{xu2024machine,chowdhury2025towards}. Exact unlearning involves retraining the model from scratch without the specified data, which is infeasible for large-scale LLMs. Approximate unlearning modifies the model parameters post hoc to minimize the influence of the targeted data. While the latter is more practical, it lacks formal guarantees and often depends on heuristic optimization.

Although several approximate unlearning algorithms have emerged, most focus solely on parameter updates without auditing and verifying whether unlearning has actually occurred during the unlearning process \citep{thaker2024position}. In practice, empirical auditing is crucial to assess the effectiveness of such algorithms \citep{wang2025tape,thudi2022necessity}. 
Current evaluation methods for data removal generally check the presence of forgotten data in the output distributions of LLMs and quantify data leakage in a metric like Forget Quality \citep{maini2024tofu,scholten2024probabilistic}. Such evaluations suffer from surface-level assessments where a model can learn to alter its output yet still retain internal representations of it. Another line of evaluations include membership inference attacks (MIAs) that employ heuristics to determine the membership of the forgotten data after unlearning (an unsuccessful MIA indicates successful data removal) \citep{kurmanji2023towards}. However, recent evidence suggests that MIAs do not work well in LLMs and often perform poorly, even near random guess performance in some cases \citep{duan2024membership}. This undermines the objective of an unlearning audit and verification. More broadly, these approaches also fail to quantify how much information was actually removed or retained.

Current audits of LLM unlearning are superficial. They typically confirm only weak unlearning \citep{jeon2024information} which checks for changes in model outputs while failing to verify the complete removal of information from internal representations. This leaves open the risk of "shallow unlearning," where models create a false sense of privacy while retaining sensitive data, falling short of the strong unlearning mandated by regulations like GDPR \citep{cooper2024machine}.
We argue for a paradigm shift from behavioral checks to principled, information-theoretic auditing. By comparing a model's representations before and after an unlearning request, we can move beyond \textit{if} a model's behavior changed to \textit{how much} information was truly erased. To achieve this, we introduce a novel framework based on Partial Information Decomposition (PID) \citep{williams2010nonnegative}, a powerful tool for dissecting shared information \citep{dissanayake2024quantifying,halder2024quantifying,liang2023quantifying,dutta2023review,dewan2024diffusion}. Our PID-based audit formalizes two key quantities: \Unlearned{} knowledge as the unique information about forgotten data successfully erased from the model and, \residual{} knowledge as the redundant information that persists, shared between the original and unlearned models, representing a potential privacy vulnerability.

This work is the first to operationalize PID for unlearning audits. Our contributions are:
\begin{enumerate}[leftmargin=*, itemsep=0em, topsep=0em]
\item We provide empirical evidence of \textbf{shallow unlearning}, showing that significant \residual{} knowledge persists even when standard metrics suggest success (\Cref{sec:shallow}).
\item We provide a formal framework to \textbf{quantify \unlearned{} and \residual{} knowledge} using the information-theoretic concepts of uniqueness and redundancy (\Cref{sec:formalizing_terms}).
\item We demonstrate the \textbf{practical utility of our audit}, showing \Residual{} Knowledge correlates strongly with adversarial vulnerability and can power a privacy-enhancing abstention mechanism (\Cref{sec:abstention}).
\end{enumerate}
Ultimately, we offer a novel, more rigorous lens for evaluating LLM unlearning, providing developers and regulators with actionable tools to ensure data is truly forgotten.

\section{Related work}
\paragraph{LLM Unlearning}
Most existing unlearning methods for LLMs follow traditional machine unlearning approaches to minimize the influence of forget samples via gradient updates during \textbf{finetuning}. The most straightforward approach employs a mixture of forgetting and retaining objectives by performing gradient ascent updates on the non-desirable sequences and regular gradient descent updates on the desirable sequences \citep{jin2024rwku,eldan2023s,jang2022knowledge}. A variant of direct finetuning is the use of Direct Preference Optimization to unlearn by framing the forget set as a preference set \citep{maini2024tofu,zhang2024negative}. 
Another line of unlearning methods involve \textbf{localized parameter modification} instead of full finetuning \citep{huu2024effects,li2024wmdp,wu2023depn,jia2024wagle}. Recent works use \textbf{auxiliary models} for unlearning by training smaller models that can help in modifying the model's responses such that the forget set responses are not generated \citep{ji2024reversing,wang2024rkld,dong2024undial}. A final line of work based on \textbf{input/output unlearning} uses prompt engineering and post-processing methods to alter model behavior on the forget set \citep{liu2024large,thaker2024guardrail,pawelczyk2023context}. Since auxiliary and input/output unlearning methods do not modify the model representations/weights, they only work at the behavior level which is often not enough to comply with regulatory frameworks like GDPR \citep{geng2025comprehensive}.
In this work, \textbf{we focus on representative, state of the art algorithms from both paradigms of LLM unlearning where model internals are modified} - finetuning and localized editing. 

\paragraph{Unlearning Evaluations}
Most work on LLM unlearning has primarily relied on output-based or task-centric evaluation metrics \citep{thaker2024position}. These include forgetting accuracy \citep{lynch2024eight}, which measures performance drop on erased data, and membership inference attacks (MIAs) to assess residual memorization \citep{shi2024muse}. For instance, some studies employ targeted or adaptive MIAs to detect whether the model still reveals traces of erased examples in generation or classification settings. Others analyze logit differences \citep{wang2025tape}, output distributions \citep{scholten2024probabilistic}, or nearest neighbor retrieval in activation spaces \citep{li2025effective} as proxies for information removal.
While these approaches offer useful empirical signals , they often conflate information deletion with performance degradation and are sensitive to attack configuration or prompt design \citep{jeon2024information}. Moreover, many are inherently model and task specific, limiting their broader application. Our work addresses these limitations by shifting the evaluation from outputs to representation-level information decomposition, offering \textbf{a more granular and interpretable account of what is forgotten and what persists}, independent of specific downstream attacks or prompts.

\section{The Issue of Shallow Unlearning}
\label{sec:shallow}
Inspired by the concept of \textit{shallow alignment} introduced by \citet{qi2024safety}, we consolidate the notion of shallow unlearning to characterize the superficial success of unlearning algorithms in LLMs. We posit that a model undergoes shallow unlearning if it merely adapts the original model's output distribution or behavior without completely removing the influence of the forget set. The model may still retain extractable information about the forgotten data, even if it no longer explicitly generates it. Ideally, in a successful unlearning scenario, no extractable information about the forget set should be identified from the model.

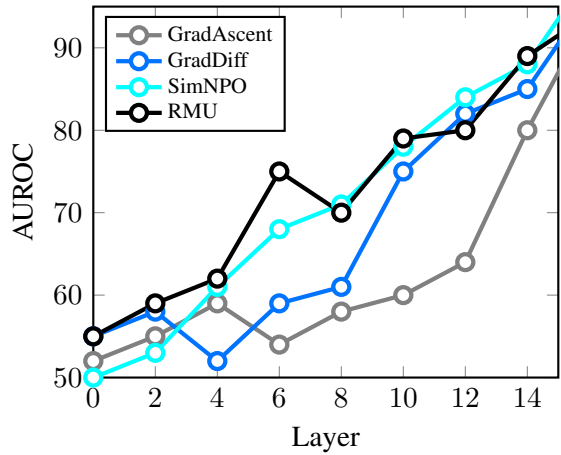
\begin{figure}[!t]
    \centering
        \begin{tikzpicture}
            \begin{axis}[
                xlabel={Layer},
                ylabel=AUROC,
                xmin=0, xmax=15,
                ymin=50, ymax=95,
                height=6.5cm,
                width=\linewidth, 
                xtick={0, 2, ..., 16},
                ytick={50,60,...,100},
                style=thick,
                ylabel near ticks,
                legend style={at={(0.025,0.975)}, anchor=north west, nodes={scale=0.75, transform shape}},
                legend cell align={left}
            ]
            \addplot+[ultra thick,mark=*,gray, mark size=3pt, mark options={scale=1, fill=white}] plot coordinates {
            (0, 52) (2, 55) (4, 59) (6, 54) (8, 58) (10, 60) (12, 64) (14, 80) (16, 94)
            };
            \addplot+[ultra thick,mark=*,brandeisblue, mark size=3pt, mark options={scale=1, fill=white}] plot coordinates {
            (0, 55) (2, 58) (4, 52) (6, 59) (8, 61) (10, 75) (12, 82) (14, 85) (16, 96)
            };
            \addplot+[ultra thick,mark=*,cyan, mark size=3pt, mark options={scale=1, fill=white}] plot coordinates {
            (0, 50) (2, 53) (4, 61) (6, 68) (8, 71) (10, 78) (12, 84) (14, 88) (16, 99)
            };
            \addplot+[ultra thick,mark=*,black, mark size=3pt, mark options={scale=1, fill=white}] plot coordinates {
            (0, 55) (2, 59) (4, 62) (6, 75) (8, 70) (10, 79) (12, 80) (14, 89) (16, 94)
            };
            \addlegendentry{GradAscent}
            \addlegendentry{GradDiff}
            \addlegendentry{SimNPO}
            \addlegendentry{RMU}
            \end{axis}
        \end{tikzpicture}
        \label{subfig:tofu}
    \hfill 
 
    \caption{Probe AUROC scores on the TOFU benchmark for different unlearning algorithms on \includegraphics[width=0.4cm]{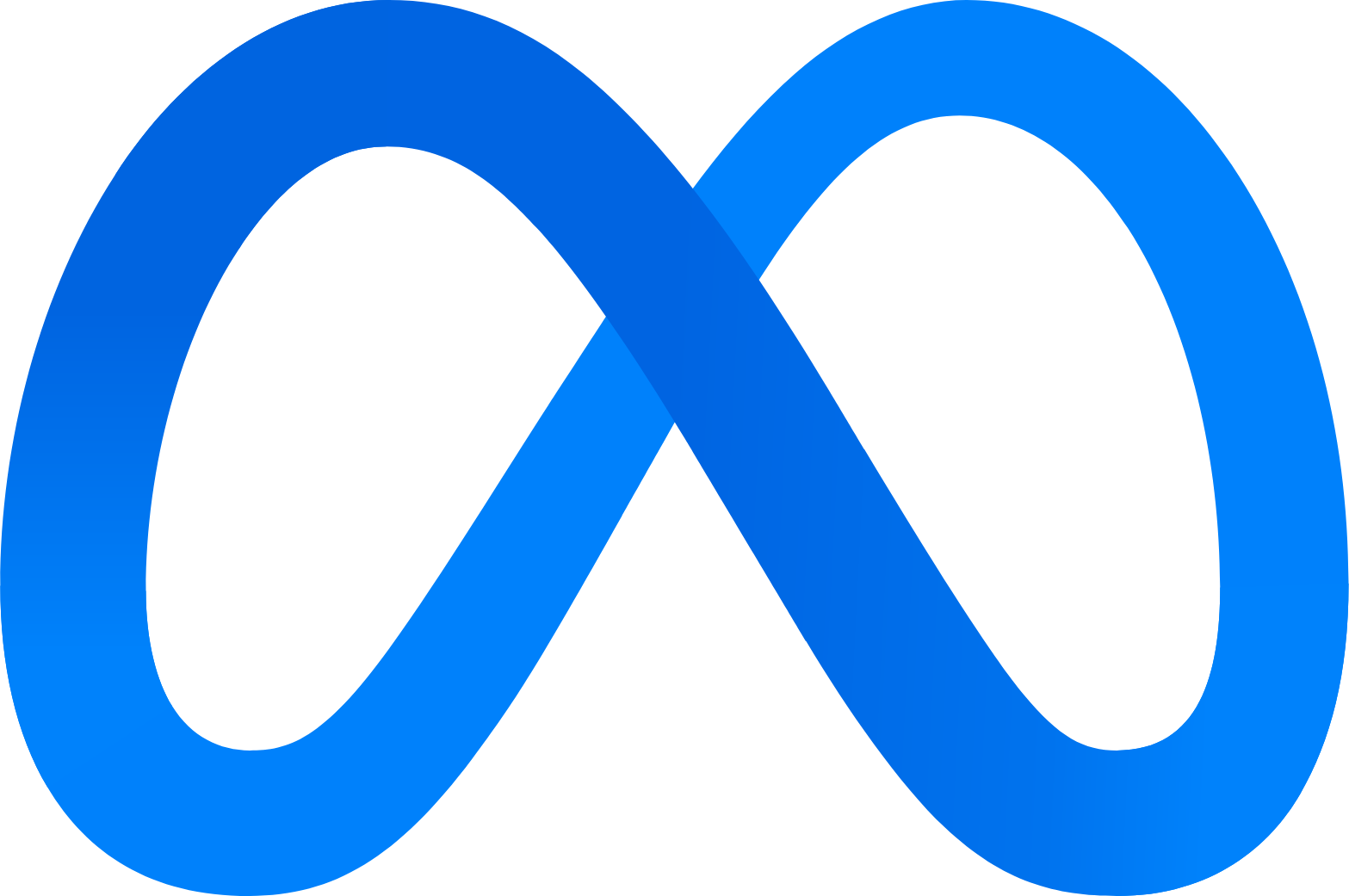}\texttt{llama}. Forget set information is linearly decodable from the residual stream of LLMs, thus undermining the sense of privacy provided by the apparent success of surface-level metrics like Forget Quality.}
    \label{fig:pid-audit-plots}
    \vspace{-10pt}
\end{figure}

To evaluate whether representations of purportedly unlearned data still retain membership information, we employ a \textit{linear probe} \citep{tenney-etal-2019-bert,alain2016understanding} trained to distinguish membership status directly from the learned model representations. Let $\mathbf{z} = f_\theta(\mathbf{x})$ denote the representation of an input $\mathbf{x}$ under the model $f_\theta$, and let $y \in \{0,1\}$ be the binary membership label indicating whether $\mathbf{x}$ was part of the data to be unlearned.
A linear probe classifier $h(\mathbf{z})$ is trained to predict $y$ from $\mathbf{z}$. The ability of this probe to reliably predict membership indicates residual information about $y$ contained in $\mathbf{z}$ despite the unlearning process. We train probes on the residual stream output from each layer of the LLM and evaluate it on similar settings as previous probing studies via measuring the AUROC score over a held-out test set \citep{kim2025linear}.

To bound the inference error of a probe, we invoke \textbf{Fano's inequality} \citep{cover1999elements}, which relates the probability of classification error to the mutual information between the inputs and the labels. Specifically, for a membership prediction task with binary labels:
\[
\mathbb{P}(h(Z) \neq Y) \geq \frac{H(Y) - I(Y; Z) - \log 2}{\log 2},
\]
where $H(Y)$ is the entropy of the membership label. This implies that if the probe can predict $y$ with low error, then the mutual information $I(Y; Z)$ must be non-trivial. Surprisingly, we find evidence of shallow unlearning (\Cref{fig:pid-audit-plots}) across all major unlearning algorithms on \texttt{llama}\footnote{\href{https://huggingface.co/meta-llama/Llama-3.2-1B-Instruct}{\includegraphics[width=0.4cm]{figures/meta-logo.png} llama-3.2-1B-Instruct}} for the TOFU \citep{maini2024tofu} and MUSE \citep{shi2024muse} unlearning benchmarks. Formally, this motivates measuring the mutual information $I(Y; Z)$, where $Z$ is the random variable over representations and $Y$ is the membership variable.

\paragraph{The Case for Information Decomposition}
While the mutual information $I(Y; Z)$ captures the total information about membership $Y$ present in representations $Z$, it does not reveal how this information compares between the base model and its unlearned counterpart - an important aspect of auditing. In particular, we aim to quantify whether the unlearning process has removed \textit{unique} information about membership that was previously encoded in the base model representations.
To this end, we apply \textit{Partial Information Decomposition (PID)} to analyze the joint informational structure of the base and unlearned representations with respect to $Y$. 

\section{Setting the Stage}
\paragraph{Notation and Problem Setting}
We consider the problem of auditing LLM unlearning from the perspective of a model deployer who has received a deletion request. 
The deployer needs to ensure that a learned model no longer retains identifiable information about a specific subset of training data after unlearning. To support empirical verification of such deletion, we aim to quantify and attribute the presence or absence of membership information in model representations before and after unlearning.

Let $Y \in \{0, 1\}$ be a binary random variable denoting the \emph{membership status} of a sample, where $Y = 1$ indicates that the sample belonged to the dataset to be forgotten. Let $B$ and $U$ denote the random variables representing the learned representations of the same input sample $\mathbf{x}$ under the base model and the unlearned model, respectively. That is,
\[
B = f_{\theta_b}(\mathbf{x}), \quad U = f_{\theta_u}(\mathbf{x}),
\]
where $f_{\theta_b}$ and $f_{\theta_u}$ are the encoding functions of the base and unlearned models at a desired layer.

Our goal is to assess whether the unlearned representation $U$ retains information about $Y$ that was present in $B$, and to what extent. This requires not only measuring the total mutual information $I(Y; B)$ and $I(Y; U)$, but also decomposing their joint contribution to $Y$ into interpretable components. To this end, we adopt a Partial Information Decomposition (PID) framework, where the joint variable $(B, U)$ serves as the set of sources and $Y$ as the target. This allows us to define and compute custom measures of unique and redundant information about membership, grounded in an operational audit framework.

\begin{figure}[t]

\begin{tikzpicture}[thick,
  set/.style = {circle, minimum size = 3cm},
  info/.style = {rounded corners=2em, draw=black, thick, inner sep=0.5cm}
]

\node[info, minimum width=7cm, minimum height=4cm, anchor=center] (box) at (0.9,0) {};

\begin{scope}
  \clip (0,0) circle(1.5cm);
  \clip (-3,-3) rectangle (0.9,3); 
  \fill[SkyBlue] (0,0) circle(1.5cm);
\end{scope}

\begin{scope}
  \clip (1.8,0) circle(1.5cm);
  \clip (0.9,-3) rectangle (4,3); 
  \fill[Orchid] (1.8,0) circle(1.5cm);
\end{scope}

\begin{scope}
  \clip (0,0) circle(1.5cm);
  \clip (1.8,0) circle(1.5cm);
  \fill[Dandelion] (0,0) circle (1.5cm);
\end{scope}

\draw (0,0) circle(1.5cm);
\draw (1.8,0) circle(1.5cm);

\node at (-1.2,1.3) {$B$};
\node at (3.0,1.3) {$U$};

\node at (-0.9,0) {$I_{\text{uniq}}^B$};
\node at (2.7,0) {$I_{\text{uniq}}^U$};
\node at (0.9,0) {$I_{\cap}$};
\node at (0.9,-1.7) {$I_{\text{syn}}$};

\end{tikzpicture}

\caption{Partial Information Decomposition. The box denotes the total joint mutual information $\mi{Y}{B,U}$, which is decomposed into four non-negative terms: synergistic information $I_\text{syn}$, redundant information \residualmath{} and the two unique information terms \unlearnedmath{} and $I_\text{uniq}^U$.}
\label{fig_pid}
\vspace{-18pt}
\end{figure}
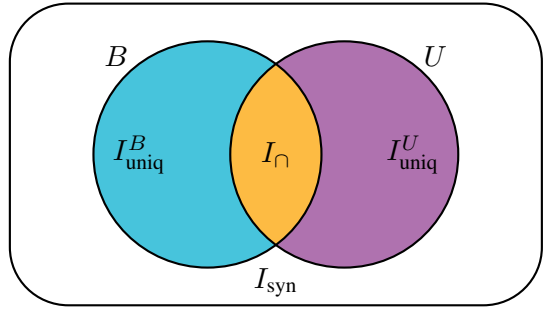

\paragraph{Background on PID:} Partial Information Decomposition (PID) \citep{williams2010nonnegative,griffith2014intersection,bertschinger2014quantifying} offers a way to decompose the joint information in two sources, say $B$ and $U$, about another random variable $Y$ (i.e., $\mi{Y}{B,U}$ where $\mi{J}{K}$ denotes the mutual information between $J$ and $K$ into four components as follows:
\begin{enumerate}[leftmargin=0.5cm, noitemsep, topsep=0pt]
    \item Unique information (\unlearnedmath{} and $I_\text{uniq}^{U}$): information about $Y$ that each source uniquely contains
    \item Redundant information (\residualmath{}): the information about $Y$ that both $B$ and $U$ share
    \item Synergistic information ($I_\text{syn}$): the information about $Y$ that can be recovered only by using both $B$ and $U$. 
\end{enumerate}
See \Cref{fig_pid} for a graphical representation. These PID components satisfy the relationships given below:
\begin{align}\label{eq:pid_terms}
    \mi{Y}{B,U} &= {\unlearnedcolor{I^B_{\text{uniq}}}} + I_\text{uniq}^{U} + {\residualcolor{I_\cap}} + I_\text{syn} \\
    \mi{Y}{B} &= {\unlearnedcolor{I^B_{\text{uniq}}}} + {\residualcolor{I_\cap}} \\
    \mi{Y}{U} &= I_\text{uniq}^{U} +  {\residualcolor{I_\cap}}
\end{align}

While this system of equations cannot be solved to arrive at a deterministic definition for each PID term, defining only one of the terms is sufficient to define the rest. 

\section{Quantifying Unlearned Information}\label{sec:formalizing_terms}

We consider the problem of unlearning in large language models (LLMs). Let $B$ denote the base model representation, $U$ the representation of the model after unlearning, and $Y$ the auditing target (e.g., a sensitive attribute or behavior). We denote by $I(Y; B)$ and $I(Y; U)$ the mutual information between $Y$ and the base or unlearned model representations, respectively. These quantities capture the extent to which the respective models retain knowledge about $Y$.

\begin{definition}[\Unlearned{} Knowledge]
\label{def_unlearnedKnowledge}
    Let $Y$ be the auditing target, and let $B$ and $U$ denote the intermediate representations of the base and unlearned models, respectively. The amount of \emph{unlearned knowledge} is defined as the information uniquely present in $B$ and absent from $U$, denoted by \unlearnedmath{}.
\end{definition}


As unlearning progresses, \unlearnedmath{} increases, while the information uniquely retained in $U$, denoted $I_\mathrm{uniq}^{U}$, decreases. By focusing specifically on \unlearnedmath{}, we isolate the quantity of interest: the information that was verifiably erased from the base model, providing a cleaner and more direct measure of unlearning efficacy.

\begin{tcolorbox}[takeaway]
    \textbf{Takeaway:} \textit{A larger \unlearnedmath{} indicates more effective unlearning.}
\end{tcolorbox}

\begin{definition}[\Residual{} Knowledge]
\label{def_residualKnowledge}
    Let $Y$ be the auditing target, and $B$, $U$ the base and unlearned model representations, respectively. The \emph{residual knowledge} shared between $B$ and $U$ is defined as \residualmath{}, representing the information about $Y$ that is still entangled between both models.
\end{definition}

Residual knowledge quantifies the shared information that persists across unlearning. This component is difficult to remove, as it reflects entangled information between $B$ and $U$. In auditing LLM unlearning, a low \residualmath{} offers stronger assurance of effective unlearning. 

\begin{tcolorbox}[takeaway]
    \textbf{Takeaway:} \textit{As \residualmath{} $\rightarrow 0$, the confidence in the unlearning process increases, potentially serving as a robust certificate of unlearning.}
\end{tcolorbox}

\subsection{Approximating Information-theoretic Terms}

While there are several methods in literature to empirically measure the redundant information \cite{bertschinger2014quantifying,williams2010nonnegative}, we use the Redundant Information Neural Estimator (RINE) \citep{e23070922} for an exact quantification. This is closer to our notion of information, reflecting the amount of information that a learned decoder can decode and reduces to several popular redundancy measures from literature (\Cref{app:pid:proof}).


\begin{definition}[Redundant information]
\label{def_redundancy}
Let $X_1, X_2$ be the source variables for the target variable $Y$ and
$f_1, f_2 \in \mathcal{V}$ be two decoders from a family of decoders
$\mathcal{V}$. Then \citet{e23070922} define a notion of redundancy as
\begin{equation}
\resizebox{\columnwidth}{!}{$
\begin{aligned}
L_{\cap}^{\mathcal{V}}(X_1; X_2 \rightarrow Y)
&:= \min_{f_1, f_2 \in \mathcal{V}} \frac{1}{2}
   \big[ H_{f_1}(Y|X_1) + H_{f_2}(Y|X_2) \big] \\
\text{s.t.}\quad
& D(f_1, f_2) = 0, \\
I_{\cap}^{\mathcal{V}}(X_1; X_2 \rightarrow Y)
&:= H(Y) - L_{\cap}^{\mathcal{V}} .
\end{aligned}
$}
\end{equation}
\end{definition}

\noindent
where $H_{f_i}(Y|X_i)$ denotes the cross-entropy when predicting $Y$ using the decoder $f_i(y|x)$ and $D(f_1,f_2) = \mathbb{E}_{x_1,x_2} [||f_1(y|x_1) - f_2(y|x_2)||_1]$ denotes the expected difference of the predictions of the two decoders. In the context of our setting, $X_1$ and $X_2$ map to $B$ and $U$ defined earlier. The model family $\mathcal{V}$ can be parameterized using simple neural networks enabling optimization over the two model families with backpropagation. A detailed explanation for the redundancy estimation is deferred to \Cref{app:pid}.

In practice, we approximate the redundancy quantities \( L_{\cap}^{\mathcal{V}} \) and \( I_{\cap}^{\mathcal{V}} \) by training two logistic regression probes \( f_1 \) and \( f_2 \), which predict the membership label \( Y \) from the respective representations of the base and unlearned models \( X_1 \) and \( X_2 \) from the final hidden layer. To enforce the agreement constraint \( D(f_1, f_2) = 0 \), we employ a Lagrangian relaxation (\Cref{eq:rine_lagrangian}). The decoders are optimized using simple backpropagation, enabling practical and efficient estimation of \( L_{\cap}^{\mathcal{V}} \) and the corresponding redundancy \( I_{\cap}^{\mathcal{V}}\). The unique information can then be naturally derived using the PID system of equations (\Cref{eq:pid_terms}).
The neural optimization used by RINE provides a principled lower bound on the true redundant information \citep{choi2024understanding}. \textbf{This property is particularly powerful in the context of an audit: a high \Residual{} Knowledge score provides a conservative but definitive signal that a non-trivial amount of information about the target at minimum still remains in the representations.} It serves as an undeniable red flag for incomplete unlearning, even if the true total amount could be higher.





\subsection{Blackwell Sufficiency \citep{blackwell1953equivalent}.}
Let $\mathcal{E}_A$ and $\mathcal{E}_B$ be two experiments observing a latent variable $Y \in \mathcal{Y}$, producing outcomes $A$ and $B$, respectively. $\mathcal{E}_A$ is said to be \textbf{Blackwell Sufficient} for $\mathcal{E}_B$ if, for any decision problem on $Y$, one can achieve at least the same expected utility with $A$ as with $B$. This holds if $B$ can be simulated from $A$ via a stochastic transformation independent of $Y$. We write this as $\mathcal{E}_A \succeq_B \mathcal{E}_B$, indicating that $A$ contains all decision-relevant information about $Y$ that $B$ does, and possibly more.

\paragraph{Application to Unlearning Audits.}
We use this framework to justify our audits. If unlearning succeeds in removing information about $Y$, then the representation $B$ should not be Blackwell Sufficient for $U$, nor vice versa, indicating informational asymmetry induced by forgetting.
Our residual knowledge measure, $I_\cap$, captures the decision-relevant information about $Y$ retained by $U$. A higher value indicates more residual knowledge and we hypothesize this represents the informational pathway that adversarial attacks exploit. Effective unlearning minimizes this redundancy. Conversely, the term $I_\text{uniq}^B$ quantifies the unique information about $Y$ that has been successfully removed.

\subsection{Experiments}\label{sec:experiments}







\begin{table*}[t!]
\centering
\resizebox{0.7\textwidth}{!}{%
\begin{tabular}{@{}llcccccc@{}}
\toprule
\textbf{Model} & \textbf{Metric} & \textbf{GA} & \textbf{GD} & \textbf{NPO} & \textbf{SimNPO} & \textbf{RMU} & \textbf{Retrained (Exact)} \\ \midrule
\multirow{3}{*}{\includegraphics[width=0.4cm]{figures/meta-logo.png} \texttt{llama}} & Forget Quality ($\uparrow$) & 0.55 & 0.62 & 0.69 & 0.73 & 0.72 & \textbf{1.00}* \\
& \Unlearned{} Knowledge ($\uparrow$) & 0.22 & 0.35 & 0.48 & 0.67 & 0.81 & -- \\
\rowcolor{Salmon!15}
\cellcolor{white} & \Residual{} Knowledge ($\downarrow$) & 0.41 & 0.32 & 0.26 & 0.14 & 0.08 & \textbf{0.002} \\ \midrule
\multirow{3}{*}{\includegraphics[width=0.4cm]{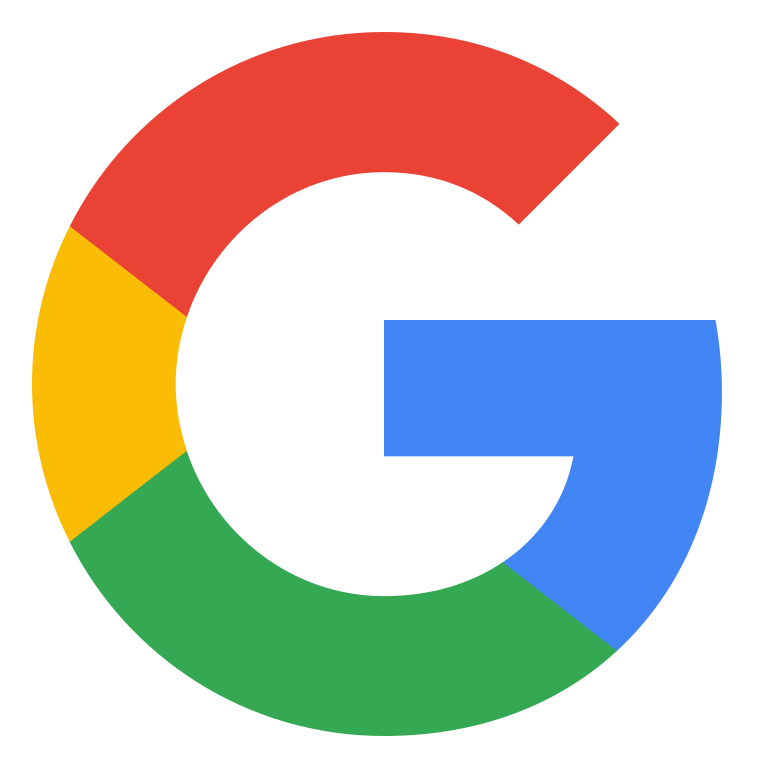}\texttt{gemma}} & Forget Quality ($\uparrow$) & 0.42 & 0.51 & 0.73 & 0.73 & 0.56 & \textbf{1.00}* \\
& \Unlearned{} Knowledge ($\uparrow$) & 0.18 & 0.31 & 0.46 & 0.61 & 0.78 & -- \\
\rowcolor{Salmon!15}
\cellcolor{white} & \Residual{} Knowledge ($\downarrow$) & 0.39 & 0.30 & 0.23 & 0.13 & 0.07 & \textbf{0.003} \\ \midrule
\multirow{3}{*}{\includegraphics[width=0.4cm]{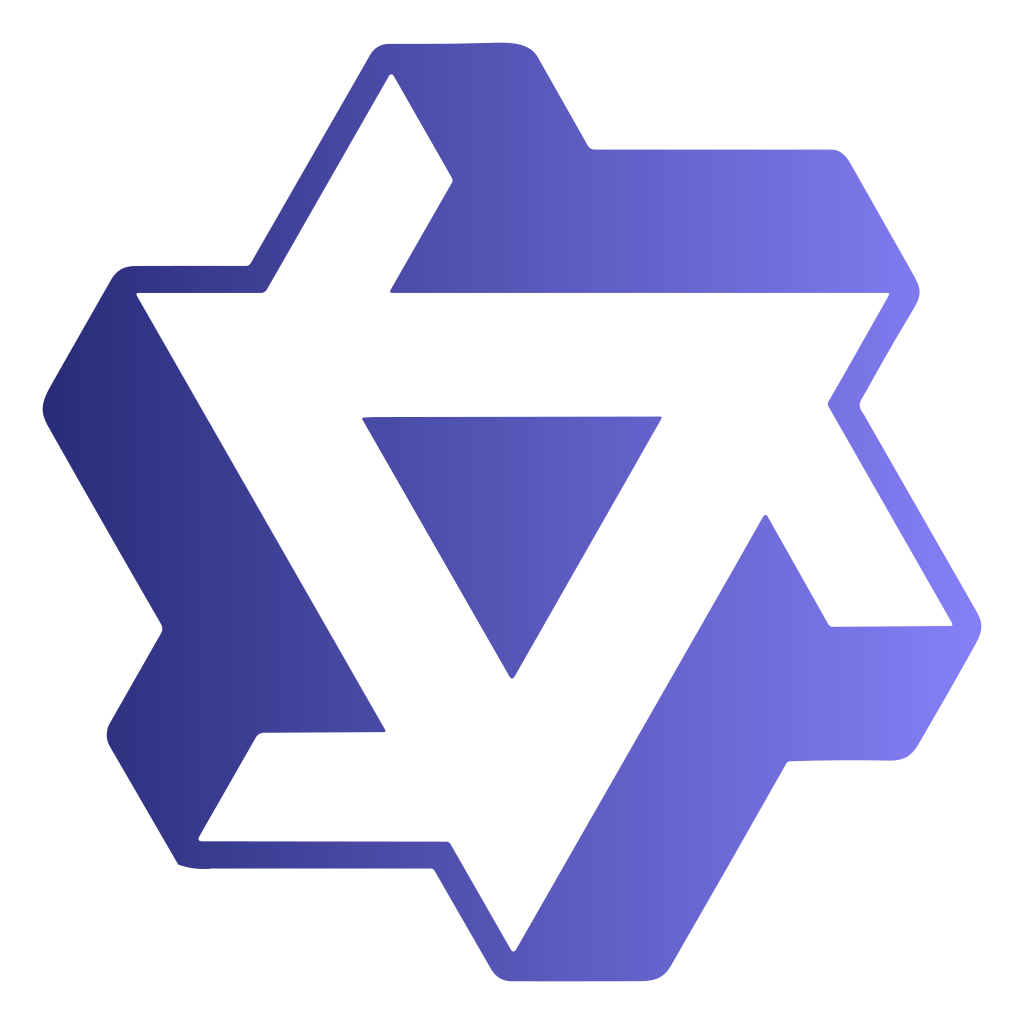} \texttt{qwen}} & Forget Quality ($\uparrow$) & 0.49 & 0.56 & 0.61 & 0.68 & 0.72 & \textbf{1.00}* \\
& \Unlearned{} Knowledge ($\uparrow$) & 0.25 & 0.36 & 0.50 & 0.69 & 0.84 & -- \\
\rowcolor{Salmon!15}
\cellcolor{white} & \Residual{} Knowledge ($\downarrow$) & 0.45 & 0.34 & 0.28 & 0.16 & 0.09 & \textbf{0.002} \\ \bottomrule
\end{tabular}%
}
\caption{Comparison of unlearning performance on the TOFU dataset across three LLMs. We report forget quality, \unlearned{} knowledge, and \residual{} knowledge. Retrained (Exact) model is the gold-standard baseline, demonstrates near-zero \residual{} knowledge, highlighting the information left behind by approximate methods. $^*$By definition, a retrained model perfectly achieves the target distribution.}\
\label{tab:pid-unlearning-comparison}
\vspace{-5pt}
\end{table*}
Now that we have established the theoretical intuition behind our proposed measures of unlearned and residual knowledge, we demonstrate experimental results to provide evidence supporting our formulation. We find that even when traditional metrics like forget quality suggest successful unlearning, our audit can provide a more granular view into the model by quantifying the amount of information that has been unlearned and information that still exists in the residual traces of the model. We use the \texttt{open-unlearning} evaluation suite to conduct our experiments.\footnote{\url{https://github.com/locuslab/open-unlearning/}}

\paragraph{Datasets.} We conduct experiments on the two LLM unlearning benchmarks: TOFU \citep{maini2024tofu} and MUSE \citep{shi2024muse}. TOFU consists of 200 fictitious author profiles split into a retain and forget set. The MUSE dataset contains BBC News articles and Harry Potter-related knowledge to unlearn. Our choice of datasets enables broad unlearning scenarios from private information to copyright content. 

\paragraph{Algorithms.} We use representative and state of the art unlearning algorithms that modify the internal structure of an LLM. In our experiments, we use Gradient Ascent (GA), Gradient Difference (GD) \citep{liu2022continual}, RMU \citep{li2024wmdp}, NPO \citep{zhang2024negative} and its variant SimNPO.

\paragraph{Models.} All experiments are conducted on three different model families: \texttt{llama} (8B) \citep{grattafiori2024llama}, \texttt{gemma} (9B) \citep{team2024gemma} and \texttt{qwen} (7B) \citep{yang2024qwen2}. We experiment with different model sizes in \Cref{apx:scalability}.

\paragraph{Results.} 
Following prior work \citep{maini2024tofu,ji2024reversing}, we report Forget Quality to measure the forget performance which evaluates how closely the unlearned LLM mirrors an LLM trained only on retain data. Table~\ref{tab:pid-unlearning-comparison} compares five unlearning algorithms across three models using both traditional metrics (forget quality) and our PID-based audit measures (\unlearned{} knowledge and \residual{} knowledge) on the TOFU dataset. Since auditing focuses on verifying successful data deletion, model utility metrics are orthogonal to the core objective and thus not central to this evaluation. Our results reveal that \textbf{Forget Quality can be misleading and may overstate the effectiveness of deletion}, \textbf{providing a false sense of privacy}.

\paragraph{Gradient-based unlearning methods retain residual knowledge.} In all three model families, the \textit{GA} and \textit{GD} baselines achieve high forget accuracy scores, suggesting successful forgetting. However, our audit shows that these methods retain substantial residual knowledge (e.g., 0.41 bits for \texttt{llama}), indicating that membership information persists in the representations even after unlearning. This residual leakage could enable a membership inference attack, contradicting the privacy guarantees implied by the accuracy metric alone.

\paragraph{Representation and preference optimization-based methods are relatively successful.} In contrast, unlearning methods such as \textit{NPO} and especially \textit{RMU} show significantly higher unlearned knowledge and lower residual knowledge, aligning more closely with the goal of irreversible data removal. \textit{RMU} consistently achieves the lowest residual knowledge across all models (e.g., 0.08 bits for \texttt{llama}), while maintaining high unlearned knowledge, indicating a meaningful reduction in identifiable traces of forgotten data.
As a crucial baseline, we include a model retrained from scratch on the retain set. As shown in \Cref{tab:pid-unlearning-comparison}, this `gold standard' model exhibits near-zero Residual Knowledge (0.002 bits), empirically validating that our metric accurately reflects the removal of information from the training process. Results on more model sizes, which show highly consistent trends, are detailed in Appendix~\ref{apx:scalability}.

Our findings are consistent across model families and show that task- and output-based metrics like forget quality miss latent internal representations. In contrast, our PID-based decomposition offers a clearer, more comprehensive view of what’s truly removed or retained after unlearning, making it a more reliable tool for privacy compliance under regulations like GDPR. 


\section{Residual Knowledge Beyond Auditing}
Beyond its role as a diagnostic quantity in our audit framework, residual knowledge serves two practical applications in the context of unlearning. 


\subsection{Residual Knowledge as a Vulnerability Indicator}\label{sec:vulnerability}


We hypothesize that models with higher \residual{} knowledge are more vulnerable to post-unlearning adversarial attacks, as more identifiable information remains in their representations. To test this, we examine the relationship between residual knowledge and attack success rates across multiple model sizes, datasets, and unlearning algorithms.
\citet{lucki2024adversarial} demonstrate that unlearning can be reversed by either finetuning on a small set of benign samples or by ablating the refusal direction of the forget preference set at inference. We apply both attacks to assess their correlation with residual knowledge.
Using attack success rates (ASR) as predictors, we fit a regression model to estimate \residual{} knowledge for models up to 7B parameters across different families. Table~\ref{tab:correlation-table} shows the resulting coefficients. In all cases, we find strong positive correlations, indicating that \residual{} knowledge is a consistent predictor of unlearning vulnerability and helps explain why adversarial attacks succeed, empirically supporting our hypothesis.


\begin{table}[t!]
\centering
\label{tab:mia_comparison}
\resizebox{\columnwidth}{!}{%
\begin{tabular}{@{}llcc@{}}
\toprule
\textbf{Model} & \textbf{Metric} & \textbf{Finetuning} & \textbf{Orthogonalization} \\ \midrule
\multirow{2}{*}{\includegraphics[width=0.4cm]{figures/meta-logo.png} \texttt{llama}} & MIA & 0.41\text{*} & 0.24\text{**} \\
 & \textbf{Our Audit} & \cellcolor{green!20}\textbf{0.60\text{**}} & \cellcolor{green!20}\textbf{0.45\text{**}} \\ \midrule
\multirow{2}{*}{\includegraphics[width=0.4cm]{figures/google-logo.png}\texttt{gemma}} & MIA & 0.39\text{**} & 0.44\text{*} \\
 & \textbf{Our Audit} & \cellcolor{green!20}\textbf{0.65\text{*}} & \cellcolor{green!20}\textbf{0.56\text{*}} \\ \midrule
\multirow{2}{*}{\includegraphics[width=0.4cm]{figures/qwen-logo.png} \texttt{qwen}} & MIA & 0.51\text{**} & 0.38\text{**} \\
 & \textbf{Our Audit} & \cellcolor{green!20}\textbf{0.71\text{**}} & \cellcolor{green!20}\textbf{0.51\text{**}} \\ \bottomrule
\end{tabular}%
}
\caption{Correlation of our \residual{} knowledge with adversarial attack success rates (ASR). We compare our \texttt{Residual Knowledge} metric against a strong MIA baseline: Min-K\%++ \citep{zhang2024min}. Our metric demonstrates a consistently stronger correlation with true model vulnerability across both attack types and model families. Significance: \text{*} p < 0.01, \text{**} p < 0.05.}
\label{tab:correlation-table}
\vspace{-15pt}
\end{table}

\begin{tcolorbox}[takeaway]
    \textbf{Takeaway:} \textit{\Residual{} knowledge can be operationalized as a lightweight auditing signal to flag insufficient unlearning and trigger remedial action or additional privacy safeguards.}
\end{tcolorbox}

\subsection{Inference-Time Abstention via Residual Risk}\label{sec:abstention}

\begin{table*}[t!]
\centering
\resizebox{\textwidth}{!}{%
\begin{tabular}{@{}llccccc@{}}
\toprule
\textbf{Sample Type} & \textbf{Query Text (Abbreviated)} & \textbf{$p_1$} & \textbf{$p_2$} & \textbf{Mean Forget Prob.} & \textbf{Disagreement} & \textbf{Risk Score} \\ \midrule
Retain Sample & Q: Yevgeny Grimkov's first work? & 0.09 & 0.12 & 0.105 & 0.97 & 0.10 \\
\rowcolor{gray!10}
Forget (Unlearned) & Q: Hsiao Yun-Hwa's popular book? & 0.95 & 0.17 & 0.560 & 0.22 & 0.12 \\
\rowcolor{red!15}
\textbf{Forget (Residual)} & \textbf{Q: Xin Lee Williams' writing style?} & \textbf{0.92} & \textbf{0.85} & \textbf{0.885} & \textbf{0.93} & \textbf{0.82} \\ \bottomrule
\end{tabular}%
}
\caption{Qualitative examples of our risk score behavior on the TOFU dataset with \includegraphics[width=0.4cm]{figures/meta-logo.png}\texttt{llama}. The risk score correctly identifies samples with residual knowledge by leveraging the confidence and agreement of membership probes on the base ($p_1$) and unlearned ($p_2$) models.}

\label{tab:qualitative_examples}
\end{table*}

Our second application explores how our framework can enable privacy-aware prediction. 
To operationalize abstention at inference time, we propose a risk score that identifies samples on which the unlearned model may still retain \residual{} information from the forget set. Let $f_1$ and $f_2$ denote the membership decoders trained on representations from the base and unlearned models, respectively, and let $p_1 = f_1(\text{forget}|x_1)$ and $p_2 = f_2(\text{forget}|x_2)$ be their predicted probabilities for the ``forget'' label. We define the risk score as:
\[
\text{RiskScore}(x) = \frac{1}{2}(p_1 + p_2) \cdot (1 - |p_1 - p_2|).
\]
This score is high when both decoders assign high probability to the sample being from the forget set \emph{and} they closely agree, indicating potential memorization that was not removed by unlearning. At inference time, we abstain from answering queries with high risk scores, thereby providing an additional safety layer over imperfect unlearning mechanisms. This approach does not require access to ground-truth membership labels at test time and is scalable via simple logistic probes trained post hoc on model representations. Refer to \Cref{app:abstention} for a detailed explanation of the abstention mechanism.
We evaluate this approach on the \texttt{llama} model under different unlearning algorithms. Table~\ref{tab:forget-retain-abstention} shows that the redundancy-based abstention rule increases unlearning performance at minimal degradation in general model utility \citep{maini2024tofu}. 

\paragraph{Comparison with Uncertainty Baselines.}
To verify that our Risk Score targets specific residual memory rather than generic model uncertainty, we compare it against a \textbf{Predictive Entropy} baseline. We compute the Shannon entropy of the next-token probability distribution, using high entropy as a proxy for uncertainty-based abstention. We evaluate this on the \texttt{llama} model unlearned via RMU. 

As shown in Table~\ref{tab:entropy_comparison}, predictive entropy yields only a modest improvement in Forget Quality ($+0.05$). in contrast, our audit-informed Risk Score achieves more than double the improvement ($+0.11$) with comparable utility cost. This empirically demonstrates that models can remain confident about forgotten data even when unlearning has failed, confirming that assessing the \textit{agreement} between base and unlearned representations is necessary to detect subtle information leakage.

\begin{table}[t]
\centering
\small
\setlength{\tabcolsep}{6pt}
\begin{tabular}{lcccc}
\toprule
\textbf{Algorithm} & \multicolumn{2}{c}{\textbf{Forget Quality}} & \multicolumn{2}{c}{\textbf{Model Utility}} \\
\cmidrule(lr){2-3} \cmidrule(lr){4-5}
& No Abst. & + Abst. & No Abst. & + Abst. \\
\midrule
GA & 0.55 & \cellcolor{green!20}\textbf{0.64} & 0.35 & 0.34 \\
GD     & 0.62 & \cellcolor{green!20}\textbf{0.71} & 0.33 & 0.31 \\
NPO   & 0.69 & \cellcolor{green!20}\textbf{0.73} & 0.43 & 0.44 \\
SimNPO     & 0.73 & \cellcolor{green!20}\textbf{0.81} & 0.54 & 0.54 \\
RMU  & 0.72 & \cellcolor{green!20}\textbf{0.83} & 0.51 & 0.49 \\
\bottomrule
\end{tabular}
\caption{Impact of abstention based on residual risk in \includegraphics[width=0.4cm]{figures/meta-logo.png} \texttt{llama} on the TOFU benchmark. Forget quality improves and general model capabilities are preserved, indicating stronger forgetting with less collateral damage.}
\label{tab:forget-retain-abstention}
\vspace{-10pt}
\end{table}

\begin{tcolorbox}[takeaway]
    \textbf{Takeaway:} \textit{By leveraging our audit-informed risk score, deployers can make fine-grained, data-driven decisions about when to trust model predictions post-unlearning.}
\end{tcolorbox}

\begin{table}[t]
\centering
\small
\resizebox{0.9\linewidth}{!}{
\begin{tabular}{lcc}
\toprule
\textbf{Method} & \textbf{Forget Quality} ($\uparrow$) & \textbf{Model Utility} ($\uparrow$) \\
\midrule
No Abstention & 0.72 & 0.51 \\
Predictive Entropy & 0.77 \small{\textcolor{gray}{(+0.05)}} & 0.48 \small{\textcolor{gray}{(-0.03)}} \\
\textbf{Our Risk Score} & \textbf{0.83} \small{\textcolor{gray}{(+0.11)}} & \textbf{0.49} \small{\textcolor{gray}{(-0.02)}} \\
\bottomrule
\end{tabular}
}
\caption{Comparison of abstention mechanisms on \includegraphics[width=0.4cm]{figures/meta-logo.png}\texttt{llama} (RMU). Our representation-based Risk Score significantly outperforms standard uncertainty (Predictive Entropy), indicating that residual knowledge is often retained with high model confidence.}
\label{tab:entropy_comparison}
\end{table}

\subsection{Qualitative Case Study: An Interpretable View of Residual Knowledge}
\label{sec:qualitative_analysis}

We examine specific examples from the TOFU dataset to illustrate how our proposed risk score distinguishes between successfully unlearned information and persistent residual knowledge. The score is designed to be high only when two conditions are met: (1) the sample is confidently identified as part of the forget set, and (2) the base and unlearned models \emph{agree} on this assessment, signaling that the unlearning process failed to remove the informational trace.

Table~\ref{tab:qualitative_examples} showcases three representative scenarios. The risk score correctly identifies high-risk samples by checking for high confidence and high agreement between the base ($p_1$) and unlearned ($p_2$) model probes.

\begin{itemize}[leftmargin=*, noitemsep, topsep=3pt]
\item \textbf{Retain/Safe}: A low forget probability ($p_1, p_2 \thickapprox 0.1$) results in a low risk score.
\item \textbf{Successfully \Unlearned{}}: High disagreement between the probes ($p_1 \gg p_2$) indicates successful information removal, leading to a low risk score.
\item \textbf{\Residual{} Knowledge}: High agreement between two confident probes ($p_1 \thickapprox p_2 \thickapprox 0.9$) signals that the informational trace persists, correctly yielding a high risk score.
\end{itemize}

\section{Discussion \& Conclusion}
We introduce a novel formalization for auditing LLM unlearning using partial information decomposition (PID) to reveal \textit{how much} and \textit{what kind of} information about erased data remains. Our \textbf{novel auditing metrics} formally decompose the unlearning process, distinguishing between information successfully removed (\unlearned{}) and information that persists (\residual{}). Unlike current unlearning assessments, our approach is interpretable, theoretically grounded and aligns with privacy goals of regulatory frameworks. We demonstrate two \textbf{practical use cases} of our audit: \textit{Vulnerability Assessment} and \textit{Abstention-based Defense}, offering actionable tools for deployers aiming to ensure privacy while preserving utility. As unlearning research advances, \textbf{standardized evaluation} is crucial. Our attack-independent, representation-level, and model-agnostic metrics lay the groundwork for formal audits and regulatory compliance for LLM unlearning. Overall, we offer a new lens on unlearning, not only as a binary success/failure criterion, but as a decomposable, quantifiable process that can be measured, improved, and deployed with confidence. 

\paragraph{The Case for White-Box Audits}

The evaluation of machine unlearning can be viewed through a lens of increasing "levels of assurance." The most basic level, achievable via black-box access, involves behavioral checks: does the model still generate the forgotten content? While useful, this provides the weakest guarantee. Our work demonstrates that models can pass such tests while still harboring "residual knowledge" internally, a phenomenon we term "shallow unlearning" (\Cref{sec:shallow}). To achieve a higher level of assurance, one must move from observing behavior to inspecting the internal mechanism. This necessitates white-box access. By directly analyzing a model's representations, our information-theoretic audit can detect these persistent informational traces that black-box methods would miss. While we employ RINE for its scalability and connection to decodability, we acknowledge the potential of alternative estimators such as the Hilbert-Schmidt Independence Criterion (HSIC)~\citep{gretton2005measuring}. We prioritized RINE because its formulation—based on optimizing families of decoders aligns conceptually with the adversarial goal of an audit: determining if information remains \textit{extractable} by a learned function. Nevertheless, comparative analysis using non-parametric estimators like HSIC remains a valuable direction for future work.

Our white-box audit is analogous to software security, where source code audits provide far stronger guarantees than black-box penetration testing alone \citep{skandylas2025automated}.
Therefore, our PID-based framework is designed for the specific, high-stakes scenario of a developer, deployer, or regulator who requires verifiable proof of data removal. \textit{In this context, white-box access is not a limitation but a prerequisite for a meaningful and trustworthy audit.}

\section*{Limitations}
We investigate the residual information with respect to data membership, future works should look into specific sensitive attritbutes relevant for deployers.
\textbf{Second}, the limitations of current LLM unlearning benchmarks carry over to our work as well: we only investigate English-language unlearning and our setting is restricted to unlearning finetuned LLMs. However, our conceptual framework should still provide valid audits for future benchmarks since our approach is data-independent and only relies on residual information encoded in the LLM representations. Additionally, the strength of our framework lies in the generalizable nature of our proposed information-theoretic measures. We only require access to two versions of the LLM. In the current study, we use the base and fine tuned model, but the core principles remain valid for other combinations an auditor might be interested in, like unlearning from instruction-tuned models, or unlearning from pre training data. 
\textbf{Third}, unlike exact unlearning, we are unable to provide certified guarantees for the unlearning process. However, we argue that our proposed measures are a practical and interpretable way to investigate unlearning in a more principled manner.
\textbf{Finally}, our framework currently audits LLM unlearning at the representation level. Our audit can be naturally extended to weights and parameters of the model which we leave as part of future work. For a black-box setting in frontier LLMs, our audit can naturally be extended assuming access only to model logits and log probabilities.

Our current work focuses on establishing a reliable framework to quantify if and how much information persists and a localisation study would be out of scope of our current work. We believe our framework lays the essential groundwork for future work to perform such fine-grained localization by applying our PID audit layer-by-layer or head-by-head. Our framework also serves as a fundamental diagnostic tool that can guide better algorithms, for example, by minimising the redundant information during unlearning.

\paragraph{On the Use of Information-Theoretic Estimators.}
Our framework relies on a neural estimator (RINE) to quantify information-theoretic measures from high-dimensional LLM representations. It is important to contextualize this choice. Accurately estimating mutual information in continuous, high-dimensional spaces is a notoriously challenging open problem. Variational estimators like RINE represent the state-of-the-art approach, but they provide a lower bound on the true information quantities.

From an auditing perspective, we argue this lower bound is not a critical weakness but a practical strength. An audit's primary goal is often risk detection. If our audit reveals a high `Residual Knowledge' score, it provides sufficient evidence to fail the unlearning process, as it proves a substantial amount of information persists. The utility of this approach is further validated by our empirical findings: `Residual Knowledge', even as a bound, correlates far more strongly with real-world adversarial vulnerability than standard MIA baselines (Table~\ref{tab:correlation-table}). While determining the precise tightness of these bounds for LLM representations remains a valuable direction for future theoretical work, our results show that the current estimators are already effective tools for practical privacy auditing.

\section*{Ethical Considerations}

This work directly engages with ethical and legal challenges surrounding user privacy in LLMs. In particular, we focus on compliance with data protection regulations such as the GDPR and the EU AI Act, which mandate the right to data deletion. Our proposed auditing framework aims to provide verifiable, interpretable evidence of data removal, enabling greater accountability in machine unlearning systems.
We emphasize that our methods are intended for responsible use by model developers, auditors, and policymakers to assess and improve privacy guarantees. However, the auditing techniques could potentially be misused to probe for residual information in models not explicitly designed for unlearning. 
All experiments in this work are conducted on publicly available benchmarks and synthetic data, without processing personally identifiable or sensitive real-world information. We advocate for future research to uphold similar ethical standards and to include auditing as a core component of privacy-preserving ML system design.


\section*{Acknowledgements}
This research work has been funded by the German Federal Ministry of Education and Research and the Hessian Ministry of Higher Education, Research, Science and the Arts within their joint support of the National Research Center for Applied Cybersecurity ATHENE.

\bibliography{custom}

@article{das2025security,
  title={Security and privacy challenges of large language models: A survey},
  author={Das, Badhan Chandra and Amini, M Hadi and Wu, Yanzhao},
  journal={ACM Computing Surveys},
  volume={57},
  number={6},
  pages={1--39},
  year={2025},
  publisher={ACM New York, NY}
}

@inproceedings{chowdhury2025towards,
  title={Towards Scalable Exact Machine Unlearning Using Parameter-Efficient Fine-Tuning},
  author={Chowdhury, Somnath Basu Roy and Choromanski, Krzysztof and Sehanobish, Arijit and Dubey, Avinava and Chaturvedi, Snigdha},
booktitle={International Conference on Learning Representations},
  year={2025},
address={Singapore}
}

@inproceedings{karamolegkou2023copyright,
    title = "Copyright Violations and Large Language Models",
    author = "Karamolegkou, Antonia  and
      Li, Jiaang  and
      Zhou, Li  and
      S{\o}gaard, Anders",
    editor = "Bouamor, Houda  and
      Pino, Juan  and
      Bali, Kalika",
    booktitle = "Proceedings of the 2023 Conference on Empirical Methods in Natural Language Processing",
    month = dec,
    year = "2023",
    address = "Singapore",
    publisher = "Association for Computational Linguistics",
    url = "https://aclanthology.org/2023.emnlp-main.458/",
    doi = "10.18653/v1/2023.emnlp-main.458",
    pages = "7403--7412",
}

@inproceedings{menta2025analyzing,
    title = "Analyzing Memorization in Large Language Models through the Lens of Model Attribution",
    author = "Menta, Tarun Ram  and
      Agrawal, Susmit  and
      Agarwal, Chirag",
    editor = "Chiruzzo, Luis  and
      Ritter, Alan  and
      Wang, Lu",
    booktitle = "Proceedings of the 2025 Conference of the Nations of the Americas Chapter of the Association for Computational Linguistics: Human Language Technologies (Volume 1: Long Papers)",
    month = apr,
    year = "2025",
    address = "Albuquerque, New Mexico",
    publisher = "Association for Computational Linguistics",
    url = "https://aclanthology.org/2025.naacl-long.535/",
    doi = "10.18653/v1/2025.naacl-long.535",
    pages = "10661--10689",
    ISBN = "979-8-89176-189-6",
}

@inproceedings{bourtoule2021machine,
  title={Machine unlearning},
  author={Bourtoule, Lucas and Chandrasekaran, Varun and Choquette-Choo, Christopher A and Jia, Hengrui and Travers, Adelin and Zhang, Baiwu and Lie, David and Papernot, Nicolas},
  booktitle={2021 IEEE symposium on security and privacy (SP)},
  pages={141--159},
  year={2021},
  organization={IEEE},
  address={Virtual}
}

@article{xu2024machine,
  author={Xu, Jie and Wu, Zihan and Wang, Cong and Jia, Xiaohua},
  journal={IEEE Transactions on Emerging Topics in Computational Intelligence}, 
  title={Machine Unlearning: Solutions and Challenges}, 
  year={2024},
  volume={8},
  number={3},
  pages={2150-2168},
  doi={10.1109/TETCI.2024.3379240}
}

@INPROCEEDINGS{thaker2024position,
  author={Thaker, Pratiksha and Hu, Shengyuan and Kale, Neil and Maurya, Yash and Wu, Zhiwei Steven and Smith, Virginia},
  booktitle={2025 IEEE Conference on Secure and Trustworthy Machine Learning (SaTML)}, 
  title={Position: LLM Unlearning Benchmarks are Weak Measures of Progress}, 
  year={2025},
  address={Copenhagen, Denmark},
  pages={520-533},
  doi={10.1109/SaTML64287.2025.00035}
}

@inproceedings{wang2025tape,
author = {Wang, Weiqi and Tian, Zhiyi and Liu, An and Yu, Shui},
title = {TAPE: Tailored Posterior Difference for Auditing of Machine Unlearning},
year = {2025},
isbn = {9798400712746},
publisher = {Association for Computing Machinery},
address = {New York, NY, USA},
url = {https://doi.org/10.1145/3696410.3714875},
doi = {10.1145/3696410.3714875},
booktitle = {Proceedings of the ACM on Web Conference 2025},
pages = {3061–3072},
numpages = {12},
location = {Sydney NSW, Australia},
series = {WWW '25}
}

@inproceedings{thudi2022necessity,
  title={On the necessity of auditable algorithmic definitions for machine unlearning},
  author={Thudi, Anvith and Jia, Hengrui and Shumailov, Ilia and Papernot, Nicolas},
  booktitle={31st USENIX security symposium (USENIX Security 22)},
  pages={4007--4022},
  year={2022},
address={Boston, Massachusetts, USA}
}

@article{lynch2024eight,
  author       = {Aengus Lynch and
                  Phillip Guo and
                  Aidan Ewart and
                  Stephen Casper and
                  Dylan Hadfield{-}Menell},
  title        = {Eight Methods to Evaluate Robust Unlearning in LLMs},
  journal      = {CoRR},
  volume       = {abs/2402.16835},
  year         = {2024},
  url          = {https://doi.org/10.48550/arXiv.2402.16835},
  doi          = {10.48550/ARXIV.2402.16835},
  eprinttype    = {arXiv},
  eprint       = {2402.16835},
  bibsource    = {dblp computer science bibliography, https://dblp.org}
}

@inproceedings{kurmanji2023towards,
  title={Towards unbounded machine unlearning},
  author={Kurmanji, Meghdad and Triantafillou, Peter and Hayes, Jamie and Triantafillou, Eleni},
  booktitle={Advances in neural information processing systems},
  volume={36},
  pages={1957--1987},
  year={2023},
address={New Orleans, Louisiana, USA}
}

@inproceedings{scholten2024probabilistic,
title={A Probabilistic Perspective on Unlearning and Alignment for Large Language Models},
author={Yan Scholten and Stephan G{\"u}nnemann and Leo Schwinn},
booktitle={The Thirteenth International Conference on Learning Representations},
year={2025},
url={https://openreview.net/forum?id=51WraMid8K},
address={Singapore}
}

@inproceedings{cooper2024machine,
  title={Machine Unlearning Doesn't Do What You Think: Lessons for Generative AI Policy, Research, and Practice},
  author={Cooper, A Feder and Choquette-Choo, Christopher A and Bogen, Miranda and Jagielski, Matthew and Filippova, Katja and Liu, Ken Ziyu and Chouldechova, Alexandra and Hayes, Jamie and Huang, Yangsibo and Mireshghallah, Niloofar and others},
  booktitle={Advances in neural information processing systems},
  year={2025},
address={San Diego, USA}
}

@inproceedings{li2025effective,
    title = "Effective Skill Unlearning through Intervention and Abstention",
    author = "Li, Yongce  and
      Sun, Chung-En  and
      Weng, Tsui-Wei",
    editor = "Chiruzzo, Luis  and
      Ritter, Alan  and
      Wang, Lu",
    booktitle = "Proceedings of the 2025 Conference of the Nations of the Americas Chapter of the Association for Computational Linguistics: Human Language Technologies (Volume 1: Long Papers)",
    month = apr,
    year = "2025",
    address = "Albuquerque, New Mexico",
    publisher = "Association for Computational Linguistics",
    url = "https://aclanthology.org/2025.naacl-long.322/",
    doi = "10.18653/v1/2025.naacl-long.322",
    pages = "6358--6371",
    ISBN = "979-8-89176-189-6"
}

@Article{williams2010nonnegative,
AUTHOR = {Mages, Tobias and Anastasiadi, Elli and Rohner, Christian},
TITLE = {Non-Negative Decomposition of Multivariate Information: From Minimum to Blackwell-Specific Information},
JOURNAL = {Entropy},
VOLUME = {26},
YEAR = {2024},
NUMBER = {5},
ARTICLE-NUMBER = {424},
URL = {https://www.mdpi.com/1099-4300/26/5/424},
PubMedID = {38785673},
ISSN = {1099-4300},
DOI = {10.3390/e26050424}
}

@article{griffith2014intersection,
  title={Intersection information based on common randomness},
  author={Griffith, Virgil and Chong, Edwin KP and James, Ryan G and Ellison, Christopher J and Crutchfield, James P},
  journal={Entropy},
  volume={16},
  number={4},
  pages={1985--2000},
  year={2014},
  publisher={Molecular Diversity Preservation International (MDPI)}
}

@article{bertschinger2014quantifying,
  title={Quantifying unique information},
  author={Bertschinger, Nils and Rauh, Johannes and Olbrich, Eckehard and Jost, J{\"u}rgen and Ay, Nihat},
  journal={Entropy},
  volume={16},
  number={4},
  pages={2161--2183},
  year={2014},
  publisher={Multidisciplinary Digital Publishing Institute}
}

@InProceedings{dissanayake2024quantifying,
  title = 	 {Quantifying Knowledge Distillation using Partial Information Decomposition},
  author =       {Dissanayake, Pasan and Hamman, Faisal and Halder, Barproda and Sucholutsky, Ilia and Zhang, Qiuyi and Dutta, Sanghamitra},
  booktitle = 	 {Proceedings of The 28th International Conference on Artificial Intelligence and Statistics},
  pages = 	 {4474--4482},
  year = 	 {2025},
  editor = 	 {Li, Yingzhen and Mandt, Stephan and Agrawal, Shipra and Khan, Emtiyaz},
  volume = 	 {258},
  series = 	 {Proceedings of Machine Learning Research},
  month = 	 {03--05 May},
  publisher =    {PMLR},
address={Mai Khao, Thailand}
}

@article{halder2024quantifying,
  title={Towards Formalizing Spuriousness of Biased Datasets Using Partial Information Decomposition},
  author={Halder, Barproda and Hamman, Faisal and Dissanayake, Pasan and Zhang, Qiuyi and Sucholutsky, Ilia and Dutta, Sanghamitra},
  journal={Transactions on machine learning research},
  year={2025},
  publisher={Transactions on Machine Learning Research (TMLR)}
}

@inproceedings{liang2023quantifying,
  title={Quantifying \& modeling multimodal interactions: An information decomposition framework},
  author={Liang, Paul Pu and Cheng, Yun and Fan, Xiang and Ling, Chun Kai and Nie, Suzanne and Chen, Richard and Deng, Zihao and Allen, Nicholas and Auerbach, Randy and Mahmood, Faisal and others},
  booktitle={Advances in Neural Information Processing Systems},
  volume={36},
  pages={27351--27393},
  year={2023},
address={New Orleans, Louisiana, USA}
}

@article{dutta2023review,
  title={A review of partial information decomposition in algorithmic fairness and explainability},
  author={Dutta, Sanghamitra and Hamman, Faisal},
  journal={Entropy},
  volume={25},
  number={5},
  pages={795},
  year={2023},
  publisher={MDPI}
}

@inproceedings{yao2024machine,
    title = "Machine Unlearning of Pre-trained Large Language Models",
    author = "Yao, Jin  and
      Chien, Eli  and
      Du, Minxin  and
      Niu, Xinyao  and
      Wang, Tianhao  and
      Cheng, Zezhou  and
      Yue, Xiang",
    editor = "Ku, Lun-Wei  and
      Martins, Andre  and
      Srikumar, Vivek",
    booktitle = "Proceedings of the 62nd Annual Meeting of the Association for Computational Linguistics (Volume 1: Long Papers)",
    month = aug,
    year = "2024",
    address = "Bangkok, Thailand",
    publisher = "Association for Computational Linguistics",
    url = "https://aclanthology.org/2024.acl-long.457/",
    doi = "10.18653/v1/2024.acl-long.457",
    pages = "8403--8419",
}

@article{zhang2024safe,
  author       = {Zhexin Zhang and
                  Junxiao Yang and
                  Pei Ke and
                  Shiyao Cui and
                  Chujie Zheng and
                  Hongning Wang and
                  Minlie Huang},
  title        = {Safe Unlearning: {A} Surprisingly Effective and Generalizable Solution
                  to Defend Against Jailbreak Attacks},
  journal      = {CoRR},
  volume       = {abs/2407.02855},
  year         = {2024},
  url          = {https://doi.org/10.48550/arXiv.2407.02855},
  doi          = {10.48550/ARXIV.2407.02855},
  eprinttype    = {arXiv},
  eprint       = {2407.02855},
  bibsource    = {dblp computer science bibliography, https://dblp.org}
}

@inproceedings{jin2024rwku,
  title={Rwku: Benchmarking real-world knowledge unlearning for large language models},
  author={Jin, Zhuoran and Cao, Pengfei and Wang, Chenhao and He, Zhitao and Yuan, Hongbang and Li, Jiachun and Chen, Yubo and Liu, Kang and Zhao, Jun},
  booktitle={Advances in Neural Information Processing Systems},
  volume={37},
  pages={98213--98263},
  year={2024},
address={Vancouver, Canada}
}

@article{eldan2023s,
  author       = {Ronen Eldan and
                  Mark Russinovich},
  title        = {Who's Harry Potter? Approximate Unlearning in LLMs},
  journal      = {CoRR},
  volume       = {abs/2310.02238},
  year         = {2023},
  url          = {https://doi.org/10.48550/arXiv.2310.02238},
  doi          = {10.48550/ARXIV.2310.02238},
  eprinttype    = {arXiv},
  eprint       = {2310.02238},
  bibsource    = {dblp computer science bibliography, https://dblp.org}
}

@inproceedings{jang2022knowledge,
    title = "Knowledge Unlearning for Mitigating Privacy Risks in Language Models",
    author = "Jang, Joel  and
      Yoon, Dongkeun  and
      Yang, Sohee  and
      Cha, Sungmin  and
      Lee, Moontae  and
      Logeswaran, Lajanugen  and
      Seo, Minjoon",
    editor = "Rogers, Anna  and
      Boyd-Graber, Jordan  and
      Okazaki, Naoaki",
    booktitle = "Proceedings of the 61st Annual Meeting of the Association for Computational Linguistics (Volume 1: Long Papers)",
    month = jul,
    year = "2023",
    address = "Toronto, Canada",
    publisher = "Association for Computational Linguistics",
    url = "https://aclanthology.org/2023.acl-long.805/",
    doi = "10.18653/v1/2023.acl-long.805",
    pages = "14389--14408",
}

@inproceedings{maini2024tofu,
  title={Tofu: A task of fictitious unlearning for llms},
  author={Maini, Pratyush and Feng, Zhili and Schwarzschild, Avi and Lipton, Zachary C and Kolter, J Zico},
  booktitle={Conference on Language Modeling},
  year={2024},
  address={Philadelphia, PA}
}

@inproceedings{zhang2024negative,
  title={Negative preference optimization: From catastrophic collapse to effective unlearning},
  author={Zhang, Ruiqi and Lin, Licong and Bai, Yu and Mei, Song},
  booktitle={Conference on Language Modeling},
  year={2024},
  address={Philadelphia, PA}
}

@InProceedings{li2024wmdp,
  title = 	 {The {WMDP} Benchmark: Measuring and Reducing Malicious Use with Unlearning},
  author =       {Li, Nathaniel and Pan, Alexander and Gopal, Anjali and Yue, Summer and Berrios, Daniel and Gatti, Alice and Li, Justin D. and Dombrowski, Ann-Kathrin and Goel, Shashwat and Mukobi, Gabriel and Helm-Burger, Nathan and Lababidi, Rassin and Justen, Lennart and Liu, Andrew Bo and Chen, Michael and Barrass, Isabelle and Zhang, Oliver and Zhu, Xiaoyuan and Tamirisa, Rishub and Bharathi, Bhrugu and Herbert-Voss, Ariel and Breuer, Cort B and Zou, Andy and Mazeika, Mantas and Wang, Zifan and Oswal, Palash and Lin, Weiran and Hunt, Adam Alfred and Tienken-Harder, Justin and Shih, Kevin Y. and Talley, Kemper and Guan, John and Steneker, Ian and Campbell, David and Jokubaitis, Brad and Basart, Steven and Fitz, Stephen and Kumaraguru, Ponnurangam and Karmakar, Kallol Krishna and Tupakula, Uday and Varadharajan, Vijay and Shoshitaishvili, Yan and Ba, Jimmy and Esvelt, Kevin M. and Wang, Alexandr and Hendrycks, Dan},
  booktitle = 	 {Proceedings of the 41st International Conference on Machine Learning},
  pages = 	 {28525--28550},
  year = 	 {2024},
  editor = 	 {Salakhutdinov, Ruslan and Kolter, Zico and Heller, Katherine and Weller, Adrian and Oliver, Nuria and Scarlett, Jonathan and Berkenkamp, Felix},
  volume = 	 {235},
  series = 	 {Proceedings of Machine Learning Research},
  month = 	 {21--27 Jul},
  publisher =    {PMLR},
}

@inproceedings{huu2024effects,
author = {Huu-Tien, Dang and Pham, Tin and Thanh-Tung, Hoang and Inoue, Naoya},
title = {On effects of steering latent representation for large language model unlearning},
year = {2025},
isbn = {978-1-57735-897-8},
publisher = {AAAI Press},
url = {https://doi.org/10.1609/aaai.v39i22.34544},
doi = {10.1609/aaai.v39i22.34544},
booktitle = {Proceedings of the Thirty-Ninth AAAI Conference on Artificial Intelligence and Thirty-Seventh Conference on Innovative Applications of Artificial Intelligence and Fifteenth Symposium on Educational Advances in Artificial Intelligence},
articleno = {2646},
numpages = {10},
series = {AAAI'25/IAAI'25/EAAI'25},
address={Philadelphia, Pennsylvania, USA}
}

@inproceedings{wu2023depn,
    title = "{DEPN}: Detecting and Editing Privacy Neurons in Pretrained Language Models",
    author = "Wu, Xinwei  and
      Li, Junzhuo  and
      Xu, Minghui  and
      Dong, Weilong  and
      Wu, Shuangzhi  and
      Bian, Chao  and
      Xiong, Deyi",
    editor = "Bouamor, Houda  and
      Pino, Juan  and
      Bali, Kalika",
    booktitle = "Proceedings of the 2023 Conference on Empirical Methods in Natural Language Processing",
    month = dec,
    year = "2023",
    address = "Singapore",
    publisher = "Association for Computational Linguistics",
    url = "https://aclanthology.org/2023.emnlp-main.174/",
    doi = "10.18653/v1/2023.emnlp-main.174",
    pages = "2875--2886",
}

@inproceedings{jia2024wagle,
  title={Wagle: Strategic weight attribution for effective and modular unlearning in large language models},
  author={Jia, Jinghan and Liu, Jiancheng and Zhang, Yihua and Ram, Parikshit and Baracaldo, Nathalie and Liu, Sijia},
  booktitle={Advances in Neural Information Processing Systems},
  volume={37},
  pages={55620--55646},
  year={2024},
address={Vancouver, Canada}
}

@inproceedings{ji2024reversing,
  title={Reversing the forget-retain objectives: An efficient llm unlearning framework from logit difference},
  author={Ji, Jiabao and Liu, Yujian and Zhang, Yang and Liu, Gaowen and Kompella, Ramana R and Liu, Sijia and Chang, Shiyu},
  booktitle={Advances in Neural Information Processing Systems},
  volume={37},
  pages={12581--12611},
  year={2024},
address={Vancouver, Canada}
}

@inproceedings{wang2024rkld,
    title = "Balancing Forget Quality and Model Utility: A Reverse {KL}-Divergence Knowledge Distillation Approach for Better Unlearning in {LLM}s",
    author = "Wang, Bichen  and
      Zi, Yuzhe  and
      Sun, Yixin  and
      Zhao, Yanyan  and
      Qin, Bing",
    editor = "Chiruzzo, Luis  and
      Ritter, Alan  and
      Wang, Lu",
    booktitle = "Proceedings of the 2025 Conference of the Nations of the Americas Chapter of the Association for Computational Linguistics: Human Language Technologies (Volume 1: Long Papers)",
    month = apr,
    year = "2025",
    address = "Albuquerque, New Mexico",
    publisher = "Association for Computational Linguistics",
    url = "https://aclanthology.org/2025.naacl-long.60/",
    doi = "10.18653/v1/2025.naacl-long.60",
    pages = "1306--1321",
}

@inproceedings{dong2024undial,
    title = "{UNDIAL}: Self-Distillation with Adjusted Logits for Robust Unlearning in Large Language Models",
    author = "Dong, Yijiang River  and
      Lin, Hongzhou  and
      Belkin, Mikhail  and
      Huerta, Ramon  and
      Vuli{\'c}, Ivan",
    editor = "Chiruzzo, Luis  and
      Ritter, Alan  and
      Wang, Lu",
    booktitle = "Proceedings of the 2025 Conference of the Nations of the Americas Chapter of the Association for Computational Linguistics: Human Language Technologies (Volume 1: Long Papers)",
    month = apr,
    year = "2025",
    address = "Albuquerque, New Mexico",
    publisher = "Association for Computational Linguistics",
    url = "https://aclanthology.org/2025.naacl-long.444/",
    doi = "10.18653/v1/2025.naacl-long.444",
    pages = "8827--8840",
}

@inproceedings{liu2024large,
  title={Large language model unlearning via embedding-corrupted prompts},
  author={Liu, Chris and Wang, Yaxuan and Flanigan, Jeffrey and Liu, Yang},
  booktitle={Advances in Neural Information Processing Systems},
  volume={37},
  pages={118198--118266},
  year={2024},
address={Vancouver, Canada}
}

@inproceedings{thaker2024guardrail,
  title={I'm not familiar with the name Harry Potter: Prompting Baselines for Unlearning in LLMs},
  author={Thaker, Pratiksha and Maurya, Yash and Smith, Virginia},
  booktitle={ICLR 2024 Workshop on Secure and Trustworthy Large Language Models},
  year={2024}
}

@inproceedings{pawelczyk2023context,
author = {Pawelczyk, Martin and Neel, Seth and Lakkaraju, Himabindu},
title = {In-context unlearning: language models as few-shot unlearners},
year = {2024},
publisher = {JMLR.org},
booktitle = {Proceedings of the 41st International Conference on Machine Learning},
articleno = {1622},
numpages = {17},
location = {Vienna, Austria},
series = {ICML'24}
}

@article{geng2025comprehensive,
  author       = {Jiahui Geng and
                  Qing Li and
                  Herbert Woisetschlaeger and
                  Zongxiong Chen and
                  Yuxia Wang and
                  Preslav Nakov and
                  Hans{-}Arno Jacobsen and
                  Fakhri Karray},
  title        = {A Comprehensive Survey of Machine Unlearning Techniques for Large
                  Language Models},
  journal      = {CoRR},
  volume       = {abs/2503.01854},
  year         = {2025},
  url          = {https://doi.org/10.48550/arXiv.2503.01854},
  doi          = {10.48550/ARXIV.2503.01854},
  eprinttype    = {arXiv},
  eprint       = {2503.01854},
  bibsource    = {dblp computer science bibliography, https://dblp.org}
}

@inproceedings{qi2024safety,
  title={Safety alignment should be made more than just a few tokens deep},
  author={Qi, Xiangyu and Panda, Ashwinee and Lyu, Kaifeng and Ma, Xiao and Roy, Subhrajit and Beirami, Ahmad and Mittal, Prateek and Henderson, Peter},
address={Singapore},
booktitle={International Conference on Learning Representations},
year = {2025}
}

@inproceedings{kim2025linear,
  title={Linear Representations of Political Perspective Emerge in Large Language Models},
  author={Kim, Junsol and Evans, James and Schein, Aaron},
address={Singapore},
booktitle={International Conference on Learning Representations},
year={2025}
}

@book{cover1999elements,
  title={Elements of information theory},
  author={Cover, Thomas M},
  year={1999},
  publisher={John Wiley \& Sons}
}

@article{blackwell1953equivalent,
  title={Equivalent comparisons of experiments},
  author={Blackwell, David},
  journal={The annals of mathematical statistics},
  pages={265--272},
  year={1953},
  publisher={JSTOR}
}

@inproceedings{goel2025differentially,
  title={Differentially Private Steering for Large Language Model Alignment},
  author={Goel, Anmol and Hu, Yaxi and Gurevych, Iryna and Sanyal, Amartya},
address={Singapore},
booktitle={International Conference on Learning Representations},
year = {2025}
}

@inproceedings{shi2024muse,
  title={Muse: Machine unlearning six-way evaluation for language models},
  author={Shi, Weijia and Lee, Jaechan and Huang, Yangsibo and Malladi, Sadhika and Zhao, Jieyu and Holtzman, Ari and Liu, Daogao and Zettlemoyer, Luke and Smith, Noah A and Zhang, Chiyuan},
address={Singapore},
booktitle={International Conference on Learning Representations},
year = {2025}
}

@inproceedings{liu2022continual,
  title={Continual learning and private unlearning},
  author={Liu, Bo and Liu, Qiang and Stone, Peter},
  booktitle={Conference on Lifelong Learning Agents},
  pages={243--254},
  year={2022},
  organization={PMLR},
address={Montreal, Canada}
}

@article{grattafiori2024llama,
  title={The llama 3 herd of models},
  author={Grattafiori, Aaron and Dubey, Abhimanyu and Jauhri, Abhinav and Pandey, Abhinav and Kadian, Abhishek and Al-Dahle, Ahmad and Letman, Aiesha and Mathur, Akhil and Schelten, Alan and Vaughan, Alex and others},
  journal={arXiv preprint arXiv:2407.21783},
  year={2024}
}

@article{team2024gemma,
  title={Gemma 2: Improving open language models at a practical size},
  author={Team, Gemma and Riviere, Morgane and Pathak, Shreya and Sessa, Pier Giuseppe and Hardin, Cassidy and Bhupatiraju, Surya and Hussenot, L{\'e}onard and Mesnard, Thomas and Shahriari, Bobak and Ram{\'e}, Alexandre and others},
  journal={arXiv preprint arXiv:2408.00118},
  year={2024}
}

@article{yang2024qwen2,
  title={Qwen2. 5 technical report},
  author={Yang, An and Yang, Baosong and Zhang, Beichen and Hui, Binyuan and Zheng, Bo and Yu, Bowen and Li, Chengyuan and Liu, Dayiheng and Huang, Fei and Wei, Haoran and others},
  journal={arXiv preprint arXiv:2412.15115},
  year={2024}
}

@article{lucki2024adversarial,
  title={An adversarial perspective on machine unlearning for ai safety},
  author={{\L}ucki, Jakub and Wei, Boyi and Huang, Yangsibo and Henderson, Peter and Tram{\`e}r, Florian and Rando, Javier},
  journal={Transactions on Machine Learning Research},
  year={2025}
}

@Article{e23070922,
AUTHOR = {Kleinman, Michael and Achille, Alessandro and Soatto, Stefano and Kao, Jonathan C.},
TITLE = {Redundant Information Neural Estimation},
JOURNAL = {Entropy},
VOLUME = {23},
YEAR = {2021},
NUMBER = {7},
ARTICLE-NUMBER = {922},
URL = {https://www.mdpi.com/1099-4300/23/7/922},
PubMedID = {34356463},
ISSN = {1099-4300},
DOI = {10.3390/e23070922}
}

@article{jeon2024information, 
title={An information theoretic evaluation metric for strong unlearning}, 
volume={40}, 
journal={Proceedings of the AAAI Conference on Artificial Intelligence}, 
author={Jeon, Dongjae and Jeung, Wonje and Kim, Taeheon and No, Albert and Choi, Jonghyun}, 
year={2025}, 
}

@inproceedings{choi2024understanding,
  title={Understanding probe behaviors through variational bounds of mutual information},
  author={Choi, Kwanghee and Jung, Jee-weon and Watanabe, Shinji},
  booktitle={ICASSP 2024-2024 IEEE International Conference on Acoustics, Speech and Signal Processing (ICASSP)},
  pages={5655--5659},
  year={2024},
  organization={IEEE},
address={Seoul, Korea}
}

@inproceedings{duan2024membership,
  title={Do Membership Inference Attacks Work on Large Language Models?}, 
  author={Michael Duan and Anshuman Suri and Niloofar Mireshghallah and Sewon Min and Weijia Shi and Luke Zettlemoyer and Yulia Tsvetkov and Yejin Choi and David Evans and Hannaneh Hajishirzi},
  year={2024},
  booktitle={Conference on Language Modeling (COLM)},
address={Philadelphia, PA, USA}
}

@inproceedings{
patil2023can,
title={Can Sensitive Information Be Deleted From {LLM}s? Objectives for Defending Against Extraction Attacks},
author={Vaidehi Patil and Peter Hase and Mohit Bansal},
booktitle={The Twelfth International Conference on Learning Representations},
year={2024},
url={https://openreview.net/forum?id=7erlRDoaV8},
address={Vienna, Austria}
}

@inproceedings{niu2024does,
  title={What does the Knowledge Neuron Thesis Have to do with Knowledge?},
  author={Niu, Jingcheng and Liu, Andrew and Zhu, Zining and Penn, Gerald},
  booktitle={International Conference on Learning Representations},
  year={2024},
address={Vienna, Austria}
}

@inproceedings{venkatesh2023gaussian,
  title={Gaussian partial information decomposition: Bias correction and application to high-dimensional data},
  author={Venkatesh, Praveen and Bennett, Corbett and Gale, Sam and Ramirez, Tamina and Heller, Greggory and Durand, Severine and Olsen, Shawn and Mihalas, Stefan},
  booktitle={Advances in Neural Information Processing Systems},
  volume={36},
  pages={74602--74635},
  year={2023},
address={New Orleans, Louisiana, USA}
}

@article{gutknecht2021bits,
  title={Bits and pieces: Understanding information decomposition from part-whole relationships and formal logic},
  author={Gutknecht, Aaron J and Wibral, Michael and Makkeh, Abdullah},
  journal={Proceedings of the Royal Society A},
  volume={477},
  number={2251},
  pages={20210110},
  year={2021},
  publisher={The Royal Society Publishing}
}

@inproceedings{dewan2024diffusion,
  title={Diffusion PID: Interpreting Diffusion via Partial Information Decomposition},
  author={Dewan, Shaurya and Zawar, Rushikesh and Saxena, Prakanshul and Chang, Yingshan and Luo, Andrew and Bisk, Yonatan},
  booktitle={Advances in Neural Information Processing Systems},
  volume={37},
  pages={2045--2079},
  year={2024},
address={Vancouver, Canada}
}

@article{griffith2015quantifying,
  title={Quantifying redundant information in predicting a target random variable},
  author={Griffith, Virgil and Ho, Tracey},
  journal={Entropy},
  volume={17},
  number={7},
  pages={4644--4653},
  year={2015},
  publisher={MDPI}
}

@article{skandylas2025automated,
  title={Automated penetration testing: Formalization and realization},
  author={Skandylas, Charilaos and Asplund, Mikael},
  journal={Computers \& Security},
  volume={155},
  pages={104454},
  year={2025},
  publisher={Elsevier}
}

@inproceedings{zhang2024min,
  title={Min-k\%++: Improved baseline for detecting pre-training data from large language models},
  author={Zhang, Jingyang and Sun, Jingwei and Yeats, Eric and Ouyang, Yang and Kuo, Martin and Zhang, Jianyi and Yang, Hao Frank and Li, Hai},
address={Singapore},
booktitle={International Conference on Learning Representations},
year = {2025}
}

@inproceedings{tenney-etal-2019-bert,
    title = "{BERT} Rediscovers the Classical {NLP} Pipeline",
    author = "Tenney, Ian  and
      Das, Dipanjan  and
      Pavlick, Ellie",
    editor = "Korhonen, Anna  and
      Traum, David  and
      M{\`a}rquez, Llu{\'i}s",
    booktitle = "Proceedings of the 57th Annual Meeting of the Association for Computational Linguistics",
    month = jul,
    year = "2019",
    address = "Florence, Italy",
    publisher = "Association for Computational Linguistics",
    url = "https://aclanthology.org/P19-1452/",
    doi = "10.18653/v1/P19-1452",
    pages = "4593--4601",
}

@inproceedings{alain2016understanding,
  title={Understanding intermediate layers using linear classifier probes},
  author={Alain, Guillaume and Bengio, Yoshua},
booktitle={International Conference on Learning Representations},
address={Toulon, France},
  year={2017}
}

@inproceedings{gretton2005measuring,
  title={Measuring statistical dependence with Hilbert-Schmidt norms},
  author={Gretton, Arthur and Bousquet, Olivier and Smola, Alex and Sch{\"o}lkopf, Bernhard},
  booktitle={International conference on algorithmic learning theory},
  pages={63--77},
  year={2005},
  organization={Springer}
}

\appendix






\clearpage

\section{Partial Information Decomposition and Redundancy Estimation}
\label{app:pid}
Information theory provides powerful tools like mutual information $I(X;Y)$ to quantify the dependency between random variables $X$ and $Y$ \citep{cover1999elements}. However, standard mutual information does not naturally describe how information about a target variable $Y$ is distributed among multiple source variables $X_1, \dots, X_n$. For instance, given two sources $X_1$ and $X_2$, the mutual information $I(X_1, X_2; Y)$ tells us the total information about $Y$ contained in $(X_1, X_2)$, but it doesn't reveal how much information is unique to $X_1$, unique to $X_2$, common to both (redundant), or only available when considering them jointly (synergistic).

Understanding this distribution of information is crucial in various domains, such as designing multi-sensor systems or analyzing neural activity recorded from different brain areas. To address this limitation, \citet{williams2010nonnegative} proposed the Partial Information Decomposition (PID), a principled framework for decomposing the total information $I(X_1, \dots, X_n; Y)$ into components reflecting unique, redundant, and synergistic contributions from the sources.

For the case of two sources $X_1$ and $X_2$ and a target $Y$, the total mutual information $I(X_1, X_2; Y)$ is decomposed as:
\begin{equation}
\begin{aligned}
I(X_1, X_2; Y) = I_\text{uniq}^{X_1} + I_\text{uniq}^{X_2} + {I}_\text{syn}(X_1, X_2; Y) + \\ I_{\cap}(X_1, X_2; Y)
\label{eq:pid_decomp}
\end{aligned}
\end{equation}
where $I_\text{uniq}^{X_i}$ represents the unique information that source $X_i$ provides about $Y$ that is not available from the other sources, $ {I}_\text{syn}(X_1, X_2; Y)$ is the synergistic information about $Y$ that is only present when $X_1$ and $X_2$ are considered jointly, and $I_{\cap}(X_1, X_2; Y)$ is the redundant information about $Y$ that is common to both $X_1$ and $X_2$.
These components relate to standard information-theoretic quantities as follows:
\begin{align}
I(X_1;Y) &= I_\text{uniq}^{X_1} + I_{\cap}(X_1, X_2; Y) \label{eq:pid_i1} \\
I(X_2; Y|X_1) &= I_\text{uniq}^{X_2} + {I}_\text{syn}(X_1, X_2; Y) \label{eq:pid_i2cond} 
\end{align}
(Note that $I(X_1, X_2; Y) = I(X_1; Y) + I(X_2; Y|X_1)$, which can be seen by summing equations \ref{eq:pid_i1} and \ref{eq:pid_i2cond}).

While the PID framework is appealing, defining and computing its constituents, particularly the redundant information $I_{\cap}$, has been challenging. Existing definitions often involve difficult optimization problems that are only feasible for discrete variables over small alphabets \cite{griffith2014intersection, bertschinger2014quantifying, e23070922, venkatesh2023gaussian, gutknecht2021bits}.

\subsection{Existing Notions of Redundant Information}
\label{app:pid:existing}

The definition of redundant information $I_{\cap}(X_1, \dots, X_n; Y)$ has been a central focus of research within the PID framework. Despite some disagreement on specific properties, several desirable characteristics are widely accepted for a notion of redundancy:
\begin{itemize}
    \item \textbf{Symmetry:} $I_{\cap}(X_1; \dots; X_n \to Y)$ is invariant to the permutation of $X_1, \dots, X_n$.
    \item \textbf{Self-redundancy:} $I_{\cap}(X_i \to Y) = I(X_i; Y)$ for any source $X_i$. The redundancy of a single source with itself is simply its mutual information with the target.
    \item \textbf{Monotonicity:} $I_{\cap}(X_1; \dots; X_n \to Y) \le I_{\cap}(X_1; \dots; X_{n-1} \to Y)$. Redundancy cannot increase by adding more sources.
\end{itemize}
Several notions satisfying these properties have been proposed. Two notable examples, formulated as optimization problems over a random variable $Q$ related to the sources and target, are:

\paragraph{Intersection Information $I^{\wedge}$} Proposed by \citet{griffith2014intersection}, $I^{\wedge}$ is defined as the maximum information that a random variable $Q$ can have about $Y$, subject to $Q$ being a deterministic function of each source $X_i$:
\begin{equation}
\begin{aligned}
I^{\wedge}(X_1; \dots; X_n \to Y) := \max_{Q} I(Y; Q) \quad \\  \text{s.t.} \quad \forall i, \exists f_i \text{ (deterministic): } Q = f_i(X_i).
\label{eq:i_wedge}
\end{aligned}
\end{equation}
This definition captures information about $Y$ that is common to all sources, as it must be extractable from each source independently via a deterministic function.

\paragraph{Common Information $I^{GH}$} A more general notion was proposed by  \citet{griffith2015quantifying}, defined as:
\begin{equation}
\begin{aligned}
I^{GH}(X_1; \dots; X_n \to Y) := \max_{Q} I(Y; Q) \quad \\  \text{s.t.} \quad \forall i, I(Y; Q|X_i) = 0.
\label{eq:i_gh}
\end{aligned}
\end{equation}
The constraint $I(Y; Q|X_i) = 0$ means that $Y \leftrightarrow X_i \leftrightarrow Q$ forms a Markov chain for each $i$. This is a weaker constraint than $Q=f_i(X_i)$, allowing $Q$ to be a stochastic function of $X_i$. $I^{GH}$ thus reflects the maximum information between $Y$ and a variable $Q$ that is dependent on each $X_i$ in a way that renders $Y$ conditionally independent of $Q$ given $X_i$.

A key limitation of these existing definitions, when computed directly via optimization over $Q$ or the joint distribution, is that they are generally only feasible for discrete variables with very small alphabets, making them impractical for high-dimensional continuous data typical in machine learning applications.

\subsection{Conceptual Grounding: Blackwell Sufficiency and Residual Knowledge}
\label{subsec:blackwell_intuition}

To justify our proposed measure of \textit{residual knowledge} in an unlearned model regarding a forgotten dataset, we draw on the concept of Blackwell Sufficiency \citep{blackwell1953equivalent}, a foundational tool for comparing the informational content of statistical experiments (in our case, model representations).

\subsection{Variational Redundant Information Neural Estimator (RINE)}
\label{app:pid:rine}

To overcome the computational challenges and enable the estimation of redundant information in high-dimensional settings, \citet{e23070922} reformulate the problem as a variational optimization over a restricted family of functions.

For two sources $X_1, X_2$ and a target $Y$, RINE is defined by an optimization problem over a chosen model family $\mathcal{V}$ of functions (e.g., neural networks), representing potential decoders $f_1, f_2$:

The core of the optimization is to find decoders $f_1, f_2 \in \mathcal{V}$ that are highly similar in their predictions about $Y$ from $X_1$ and $X_2$ respectively, while minimizing their prediction error on $Y$.
Let $f_i(y|x_i)$ denote the conditional probability distribution over $Y$ given $X_i=x_i$ as modeled by decoder $f_i$.
RINE defines a loss function $\mathcal{L}_{\mathcal{V}}(X_1; X_2 \to Y)$ that seeks functions $f_1, f_2 \in \mathcal{V}$ minimizing the average prediction loss (cross-entropy) while being constrained to produce similar predictions:
\begin{equation}
\begin{aligned}
\mathcal{L}_{\mathcal{V}}(X_1; X_2 \to Y)
&:= \min_{f_1, f_2 \in \mathcal{V}}
\left[
\begin{aligned}
&\nicefrac{1}{2} H_{f_1}(Y|X_1) \\
&+ \nicefrac{1}{2} H_{f_2}(Y|X_2)
\end{aligned}
\right] \\
&\text{s.t.}\quad D(f_1, f_2) = 0
\end{aligned}
\label{eq:rine_constraint}
\end{equation}

where $H_f(Y|X) = \mathbb{E}_{(x,y) \sim p(x,y)} [-\log f(y|x)]$ is the expected cross-entropy when predicting $Y$ using decoder $f$, and $D(f_1, f_2) = \mathbb{E}_{x_1,x_2 \sim p(x_1,x_2)} [\|f_1(Y|x_1) - f_2(Y|x_2)\|_1]$ is the expected difference between the output distributions of the two decoders (e.g., measured by L1 norm, though other divergence measures could be used). The $\nicefrac{1}{2}$ terms are included for symmetry. The constraint $D(f_1, f_2) = 0$ implies that $f_1(Y|X_1)$ and $f_2(Y|X_2)$ are equal in expectation over the data distribution. If $f_1$ and $f_2$ are deterministic functions, this constraint means $f_1(X_1) = f_2(X_2)$ almost surely. If they are stochastic functions, it means their output distributions are the same.

The $\mathcal{V}$-redundant information, denoted $I_{\mathcal{V}}(X_1, X_2; Y)$, is then defined based on this minimized loss:
\begin{equation}
I_{\mathcal{V}}(X_1, X_2; Y) := H(Y) - \mathcal{L}_{\mathcal{V}}(X_1; X_2 \to Y)
\label{eq:rine_def}
\end{equation}
where $H(Y)$ is the entropy of the target variable. This definition is analogous to the definition of standard mutual information $I(X;Y) = H(Y) - H(Y|X)$, where $H(Y|X)$ is the minimum possible cross-entropy when predicting $Y$ from $X$ using \textbf{any} function. Here, we minimize cross-entropy using functions from $\mathcal{V}$ subject to the similarity constraint, effectively isolating the information about $Y$ that is extractable by similar functions from both sources.

To solve the constrained minimization problem subject to (\ref{eq:rine_constraint}) in practice, we can minimize the corresponding Lagrangian:
\begin{equation}
\begin{aligned}
\mathcal{L}_{\mathcal{V}}(X_1; X_2 \to Y, \beta)
&:= \min_{f_1, f_2 \in \mathcal{V}}
\left[
\begin{aligned}
&\nicefrac{1}{2} H_{f_1}(Y|X_1) \\
&+ \nicefrac{1}{2} H_{f_2}(Y|X_2) \\
&+ \beta D(f_1, f_2)
\end{aligned}
\right]
\end{aligned}
\label{eq:rine_lagrangian}
\end{equation}

where $\beta \ge 0$ is a Lagrange multiplier. As $\beta \to \infty$, the solution to the Lagrangian approaches the solution to the constrained problem, forcing $D(f_1, f_2) \to 0$.

A key insight is that by choosing the function family $\mathcal{V}$ appropriately, RINE generalizes the existing notions of redundancy $I^{\wedge}$ and $I^{GH}$:

\paragraph{Proposition 1} Let $\mathcal{V}$ be the family of deterministic functions. Then, $I_{\mathcal{V}}(X_1, X_2; Y) = I^{\wedge}(X_1, X_2; Y)$. If, instead, $\mathcal{V}$ is the family of stochastic functions, then $I_{\mathcal{V}}(X_1, X_2; Y) = I^{GH}(X_1, X_2; Y)$.

This proposition is formally proven below. This connection shows that optimizing over a family of functions, as done in RINE, provides a unified framework that recovers established definitions of redundancy when the function family is sufficiently rich. By using restricted families of functions like neural networks, we can approximate this redundancy in high-dimensional settings where exact computation is intractable. The optimization is performed over the parameters of the functions $f_1, f_2 \in \mathcal{V}$ (e.g., neural network weights) using gradient descent.

\subsection{Proof of Proposition 1}
\label{app:pid:proof}

We provide an intuitive proof of Proposition 1, showing how the RINE objective recovers known definitions of redundant information depending on the type of functions allowed. We borrow the proofs directly from \citet{e23070922} but provide an intuitive explanation below.

Recall the RINE loss:
\begin{equation}
\begin{aligned}
\mathcal{L}_{\mathcal{V}}(X_1; X_2 \to Y)
&:= \min_{f_1, f_2 \in \mathcal{V}}
\left[
\begin{aligned}
&\nicefrac{1}{2} H_{f_1}(Y|X_1) \\
&+ \nicefrac{1}{2} H_{f_2}(Y|X_2)
\end{aligned}
\right]
\end{aligned}
\end{equation}

subject to the constraint that the outputs of $f_1$ and $f_2$ are identical:
\begin{equation}
D(f_1, f_2) := \mathbb{E}_{x_1,x_2} \left[ \|f_1(Y|x_1) - f_2(Y|x_2)\|_1 \right] = 0
\end{equation}

This means both functions must output the same predictive distribution over $Y$, even though they receive different inputs ($X_1$ and $X_2$ respectively). Let this shared output be a variable $Q$.

Then the loss becomes:
\begin{equation}
\mathcal{L}_{\mathcal{V}}(X_1; X_2 \to Y) = H(Y|Q)
\end{equation}
and the estimated redundancy is:
\begin{equation}
I_{\mathcal{V}}(X_1, X_2; Y) = H(Y) - H(Y|Q) = I(Y; Q)
\end{equation}
So the amount of redundant information depends on how much $Q$ tells us about $Y$, where $Q$ must be derivable from both $X_1$ and $X_2$ through functions in $\mathcal{V}$.

\paragraph{Case 1: Deterministic Functions.}

Suppose $\mathcal{V}$ is the set of all deterministic functions. Then $Q$ must be a deterministic function of each input:
\[
Q = f_1(X_1) = f_2(X_2)
\]
This matches the definition of $I^{\wedge}$:
\begin{equation}
I^{\wedge}(X_1, X_2 \to Y) = \max_{Q: Q = f_i(X_i)} I(Y; Q)
\end{equation}
Therefore, under deterministic $\mathcal{V}$, the RINE estimate recovers $I^{\wedge}$.

\paragraph{Case 2: Stochastic Functions.}

Now let $\mathcal{V}$ be the set of stochastic functions. Then $Q$ can be a stochastic function of $X_1$ and $X_2$, and the constraint $D(f_1, f_2) = 0$ still requires that both $X_1$ and $X_2$ can generate the same distribution over $Q$.

This corresponds to the constraint:
\[
I(Y; Q | X_1) = 0 \quad \text{and} \quad I(Y; Q | X_2) = 0
\]
which means $Y \leftrightarrow X_i \leftrightarrow Q$ forms a Markov chain for each $i$. That is the definition of $I^{GH}$:
\begin{equation}
I^{GH}(X_1, X_2 \to Y) = \max_{Q: I(Y; Q|X_i)=0} I(Y; Q)
\end{equation}
So when $\mathcal{V}$ allows stochastic mappings, RINE recovers $I^{GH}$.

\paragraph{Conclusion.}

RINE unifies both $I^{\wedge}$ and $I^{GH}$ under a single optimization framework. By choosing different function families $\mathcal{V}$ (deterministic or stochastic), the method recovers different classical definitions of redundant information.

\section{Inference-Time Abstention via Risk Scoring}
\label{app:abstention}

In this section, we provide a detailed explanation of our inference-time abstention mechanism designed to mitigate residual privacy risks in unlearned models. Even after training-time unlearning procedures are applied, models may retain partial information from forgotten samples. Our approach provides an additional safety barrier by abstaining from answering queries that are likely to belong to the forget set and are insufficiently unlearned.

\subsection{Motivation}
Two natural candidates for an inference-time abstention score are the  \textit{redundant information} or \textit{disagreement} between the base and unlearned models to identify forgotten examples. However, these methods have key limitations:

\begin{itemize}
    \item \textbf{Redundancy-based scoring}: Redundancy metrics, such as $I_{\cap}^{\mathcal{V}}(X_1; X_2 \rightarrow Y)$, require access to ground-truth membership labels $Y$ for every test sample, which are unavailable in real-world inference settings.
    \item \textbf{Disagreement-based scoring}: Simple disagreement metrics like $\lvert f_1(y|x_1) - f_2(y|x_2) \rvert$ may misclassify successfully forgotten samples as risky (due to decoder divergence) and benign retained samples as safe (due to decoder agreement), thus leading to high false positives and negatives.
\end{itemize}

To address these issues, we propose a more targeted risk scoring mechanism that accounts for both the confidence and agreement of decoders specifically on the ``forget'' class.

\subsection{Proposed Risk Score and Abstention Algorithm}\label{app:risk_score_explained}
Let $f_1$ and $f_2$ be binary membership decoders trained on representations $x_1$ and $x_2$ from the base and unlearned models, respectively. Let:
\[
p_1 = f_1(\text{forget}|x_1), \quad p_2 = f_2(\text{forget}|x_2).
\]
We define the risk score as:
\[
\text{RiskScore}(x) = \underbrace{\frac{1}{2}(p_1 + p_2)}_{\text{Mean Forget Probability}} \cdot \underbrace{(1 - |p_1 - p_2|)}_{\text{Agreement Term}}.
\]
This risk score is high when both decoders assign high probability to the sample being from the forget set \emph{and} they closely agree on this assessment, suggesting residual memorization. \footnote{We evaluate the performance on different thresholds and observe best performance at $0.48$. This is a parameter deployers can optimize depending on their privacy-compliance.}

\begin{algorithm}[H]
\caption{Inference-Time Abstention via Risk Scoring}
\label{alg:abstention}
\begin{algorithmic}[1]
\State Trained decoders $f_1, f_2$, representations $x_1, x_2$ of input $x$, abstention threshold $\tau$
\State $p_1 \leftarrow f_1(\text{forget}|x_1)$
\State $p_2 \leftarrow f_2(\text{forget}|x_2)$
\State $\text{mean\_forget} \leftarrow 0.5 \cdot (p_1 + p_2)$
\State $\text{agreement} \leftarrow 1 - |p_1 - p_2|$
\State $\text{risk\_score} \leftarrow \text{mean\_forget} \cdot \text{agreement}$ 
\If{$\text{risk\_score} > \tau$} 
\State \textbf{Abstain}
\Else
\State \textbf{Answer}
\EndIf
\end{algorithmic}
\end{algorithm}

\begin{table*}[t]
\centering
\caption{Example of risk score behavior for different sample types.}
\label{tab:risk_examples}
\begin{tabular}{lcccccc}
\toprule
\textbf{Sample Type} & $p_1$ & $p_2$ & Mean Forget & Disagreement & Risk Score & Action \\
\midrule
Retain Sample         & 0.1   & 0.1   & 0.1         & 0.0           & 0.10       & Answer \\
Forget (Unlearned)    & 0.9   & 0.1   & 0.5         & 0.8           & 0.10       & Answer \\
Forget (Residual)     & 0.9   & 0.8   & 0.85        & 0.1           & 0.77       & Abstain \\
\bottomrule
\end{tabular}
\end{table*}

\subsection{Discussion}

The final abstention score balances two intuitions:
\begin{itemize}
    \item \textbf{High average forget probability} suggests that the sample looks like a forgotten point.
    \item \textbf{Low disagreement} indicates the unlearned model still agrees with the base model, implying forgetting may have failed.
\end{itemize}
\Cref{tab:risk_examples} provides a conceptual example to clarify the behavior of the proposed risk score across different possible types of samples. Only when both of these hold do we trigger abstention, which makes the mechanism robust against false positives on benign queries and false negatives on unsuccessfully forgotten ones. This strategy also avoids requiring access to ground-truth labels or expensive PID optimization at inference time.

\section{Scalability Experiments with Smaller Models}
\label{apx:scalability}

To verify that our auditing framework is effective across different model scales, we conducted additional experiments on smaller language models: \texttt{llama-3-1B-Instruct} and \texttt{gemma-2B}. The primary goal was to assess whether the trends observed in the 7B-9B parameter models hold true for smaller, more accessible model sizes.

As shown in Table~\ref{tab:smaller_model_results}, the results are highly consistent with our main findings. The relative performance ranking of the unlearning algorithms remains largely stable. For instance, RMU continues to be one of the most effective methods, achieving the lowest `Residual Knowledge` and highest `Unlearned Knowledge`. Conversely, simpler methods like Gradient Ascent (GA) still leave behind significant residual information, even in these smaller models.

This consistency across scales (from 1B to 9B parameters) suggests that the phenomena of "shallow unlearning" and the utility of our PID-based audit are fundamental properties of the unlearning process in current LLMs, rather than artifacts of a specific model size.

\begin{table*}[!ht]
\centering
\caption{Unlearning performance audit on smaller models (\texttt{llama-3-1B} and \texttt{gemma-2B}) using the TOFU dataset. Results show consistent trends with the larger models presented in the main paper (Table~\ref{tab:pid-unlearning-comparison}), demonstrating the scalability of our auditing framework.}
\label{tab:smaller_model_results}
\resizebox{0.7\textwidth}{!}{%
\begin{tabular}{@{}llccccc@{}}
\toprule
\textbf{Model} & \textbf{Metric} & \textbf{GA} & \textbf{GD} & \textbf{NPO} & \textbf{SimNPO} & \textbf{RMU} \\ \midrule
\multirow{3}{*}{Llama-3 1B} & Forget Quality ($\uparrow$) & 0.48 & 0.55 & 0.61 & 0.65 & 0.68 \\
& Unlearned Knowledge ($\uparrow$) & 0.31 & 0.36 & 0.51 & 0.62 & 0.80 \\
\rowcolor{red!15}
& Residual Knowledge ($\downarrow$) & 0.29 & 0.24 & 0.19 & 0.18 & 0.08 \\ \midrule
\multirow{3}{*}{Gemma 2B} & Forget Quality ($\uparrow$) & 0.45 & 0.53 & 0.68 & 0.70 & 0.59 \\
& Unlearned Knowledge ($\uparrow$) & 0.20 & 0.29 & 0.48 & 0.64 & 0.80 \\
\rowcolor{red!15}
& Residual Knowledge ($\downarrow$) & 0.33 & 0.29 & 0.18 & 0.15 & 0.09 \\ \bottomrule
\end{tabular}%
}
\end{table*}

\section{Experimental Details}
In all experiments, we use the \texttt{open-unlearning} library for access to the unlearning algorithms and benchmarks. We train simple logistic regression models for probing experiments. For redundancy computation, we train linear probes on base and unlearned model representations from the final layer, using Adam optimization in PyTorch. All experiments are conducted on NVIDIA A100 80GB GPU. For the adversarial attacks, we reuse the codebase from \citet{lucki2024adversarial} and compute ASRs on each model family (all models upto 7B parameters) and dataset combination, then run the regression reported in \Cref{sec:vulnerability}. Our framework can be applied to any unlearning algorithm and model as long as we have white-box access.

The threshold for abstention-based experiments was determined by performing a sweep on a held-out validation set to balance forget quality improvement and utility preservation. This  can be adapted by deployers based on their specific privacy-utility trade-offs.

\begin{table*}[t]
\centering
\small
\setlength{\tabcolsep}{7pt}
\begin{tabular}{lcccccc}
\toprule
\textbf{Model} & \textbf{Metric} & \textbf{GA} & \textbf{GD} & \textbf{NPO} & \textbf{SimNPO} & \textbf{RMU} \\
\midrule

\multirow{3}{*}{\texttt{llama}}
 & Forget Quality ($\uparrow$) & 0.53 & 0.60 & 0.70 & 0.74 & 0.71 \\
 & Unlearned Knowledge ($\uparrow$) & 0.24 & 0.34 & 0.47 & 0.65 & 0.80 \\
 & Residual Knowledge ($\downarrow$) & 0.42 & 0.33 & 0.27 & 0.15 & 0.09 \\

\midrule

\multirow{3}{*}{\texttt{gemma}}
 & Forget Quality ($\uparrow$) & 0.44 & 0.50 & 0.71 & 0.74 & 0.57 \\
 & Unlearned Knowledge ($\uparrow$) & 0.19 & 0.29 & 0.45 & 0.60 & 0.76 \\
 & Residual Knowledge ($\downarrow$) & 0.40 & 0.29 & 0.24 & 0.12 & 0.06 \\

\midrule

\multirow{3}{*}{\texttt{qwen}}
 & Forget Quality ($\uparrow$) & 0.48 & 0.57 & 0.63 & 0.66 & 0.73 \\
 & Unlearned Knowledge ($\uparrow$) & 0.26 & 0.38 & 0.52 & 0.68 & 0.83 \\
 & Residual Knowledge ($\downarrow$) & 0.44 & 0.35 & 0.29 & 0.17 & 0.08 \\

\bottomrule
\end{tabular}
\caption{Audit results on the MUSE dataset.}
\label{tab:pid-unlearning-comparison-randomized}
\vspace{-10pt}
\end{table*}

\end{document}